\RequirePackage{fix-cm}
\documentclass[twocolumn]{svjour3}          
\smartqed  
\usepackage{hyperref}
\usepackage{tabu}
\usepackage{tabularx}
\usepackage{epsfig}
\usepackage{graphicx}
\usepackage{amsmath}
\usepackage{amssymb}
\usepackage{subfigure}
\usepackage{enumitem}
\usepackage{tabu}
\usepackage{tabularx}
\usepackage{booktabs}
\usepackage{diagbox}
\usepackage{array}
\usepackage{float}
\usepackage{threeparttable}
\usepackage{multirow} 
\usepackage{dsfont}
\usepackage{amsmath,amsfonts}
\usepackage{algorithmic}
\usepackage{algorithm}
\usepackage{array}
\usepackage{textcomp}
\usepackage{stfloats}
\usepackage{url}
\usepackage{verbatim}
\usepackage{graphicx}
\usepackage{cite}
\usepackage[normalem]{ulem}
\usepackage{color, colortbl}
\usepackage{bm}
\usepackage{array}

\usepackage{microtype}
\usepackage{graphicx}
\usepackage{cuted} 
\usepackage{capt-of}
\PassOptionsToPackage{table}{xcolor} 
\definecolor{mygray}{gray}{.90}
\definecolor{reda}{RGB}{202,0,0}
\definecolor{redb}{RGB}{217,148,143}
\definecolor{myyellow}{RGB}{190,144,0}
\definecolor{mygreen}{RGB}{0,136,51}
\definecolor{myblue}{RGB}{0,102,204}
\usepackage{colortbl}

\begin{document}

\title{Power Battery Detection}

\titlerunning{Power Battery Detection}        

\author{Xiaoqi Zhao$^{1}$ \and
	    Peiqian Cao$^{2}$ \and
	    Chenyang Yu$^{2}$ \and
	    Zonglei Feng$^{3}$ \and \\
	    Lihe Zhang$^{2}$ \and
	    Hanqi Liu$^{4}$ \and
	    Jiaming Zuo$^{4}$ \and 
        Youwei Pang$^{5}$  \and \\
        Jinsong Ouyang$^{1}$ \and
        Weisi Lin$^{5}$ \and
        Georges El Fakhri$^{1}$ \and\\
        Huchuan Lu$^{2}$ \and
        Xiaofeng Liu$^{1}$ \\
        \\
         {\tt \url{https://github.com/NTU-AI4X/X-ray-PBD}}
}


\institute{
$^{1}$  Yale University, USA\\
$^{2}$  Dalian University of Technology, China\\
$^{3}$  Volkswagen Automotive Co., Ltd \\
$^{4}$  X3000 Inspection Co., Ltd\\
$^{5}$  Nanyang Technological University, Singapore\\
Lihe~Zhang and Youwei~Pang are the corresponding authors.
\\
\\
E-mail: \\
xiaoqi.zhao@yale.edu \\
zhaoxq.cv@gmail.com \\
zhanglihe@dlut.edu.cn \\
1artpang@gmail.com \\
xiaofeng.liu@yale.edu\\
\\
}

\date{Received: date / Accepted: date}

\maketitle
\begin{abstract}
Power batteries are essential components in electric vehicles, where internal structural defects can pose serious safety risks. We conduct a comprehensive study on a new task, power battery detection (PBD), which aims to localize the dense endpoints of cathode and anode plates from industrial X-ray images for quality inspection. Manual inspection is inefficient and error-prone, while traditional vision algorithms struggle with densely packed plates, low contrast, scale variation, and imaging artifacts.
 To address this issue and drive more attention into this meaningful task, we present PBD5K, the first large-scale benchmark for this task, consisting of 5,000 X-ray images from nine battery types with fine-grained annotations and eight types of real-world visual interference. To support scalable and consistent labeling, we develop an intelligent annotation pipeline that combines image filtering, model-assisted pre-labeling, cross-verification, and layered quality evaluation.
We formulate PBD as a point-level segmentation problem and propose MDCNeXt, a model designed to extract and integrate multi-dimensional structure clues including point, line, and count information from the plate itself. To improve discrimination between plates and suppress visual interference, MDCNeXt incorporates two state space modules. The first is a prompt-filtered module that learns contrastive relationships guided by task-specific prompts. The second is a density-aware reordering module that refines segmentation in regions with high plate density.  In addition, we propose a distance-adaptive mask generation strategy to provide robust supervision under varying spatial distributions of anode and cathode positions. 
Without any bells and whistles, our segmentation-based  MDCNeXt consistently outperforms various other corner detection,
 crowd counting and general/tiny object detection-based solutions, making it a strong baseline that can help facilitate
 future research in PBD. 
 Finally, we provide a detailed discussion of future directions and challenges for research in power battery detection. 
\end{abstract}

\begin{figure*}[!t]
	\centering
	\includegraphics[width=\linewidth]{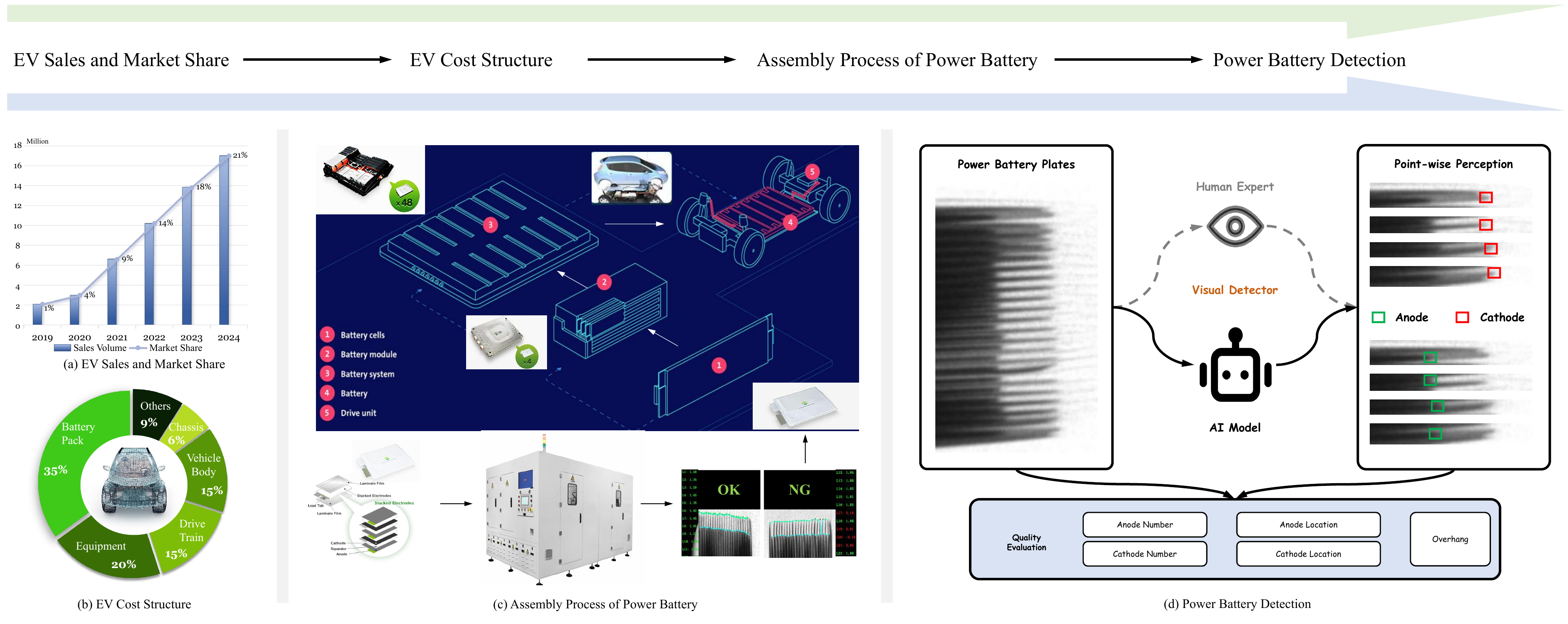}
	\caption{Motivation of power battery detection (PBD). (a) EV sales are rapidly increasing, raising battery safety demands. (b) Battery packs account for $\sim$35\% of EV cost, highlighting their critical role. (c) Assembly process of power batteries for EVs, where PBD is applied to the battery cells before assembly. (d) Illustration of the power battery detection task.
    }
	 \label{fig:development_EV}
\end{figure*}

\section{Introduction}
With the rapid adoption of electric vehicles (EVs), Fig.~\ref{fig:development_EV}(a) shows that EV market share has grown from around 1\% to over 20\% in the past five years, with 17.1 million units sold globally in 2024~\cite{evsales2024,ieaev2024}.
The power battery serves as the sole energy source for EVs, directly influencing their performance, range, and safety~\cite{BEV_1}.
As shown in Fig.~\ref{fig:development_EV}(b), the battery pack contributes nearly 35\% of total manufacturing cost, underscoring its critical economic and functional role. Ensuring its structural integrity requires reliable inspection throughout production.
During cell assembly (Fig.~\ref{fig:development_EV}(c)), hundreds of cathode and anode plates are stacked alternately. Any misalignment or overhang can lead to short circuits, overheating, or even explosion, creating serious safety risks.
To address this, a post-assembly inspection process, termed power battery detection (PBD), is widely applied.
As shown in Fig.~\ref{fig:development_EV}(d), PBD uses X-ray images captured by digital radiography (DR) systems to localize cathode and anode endpoints with coordinate-level precision.
This enables automatic computation of key quality indicators, such as the number of plates and their overhang\footnote{Number and overhang are two critical metrics in battery quality control. Number indicates the total count of anode/cathode plates, while overhang measures the average vertical offset between one anode and its adjacent cathodes.}, which determine whether a battery cell is acceptable (OK) or defective (No Good, NG).
However, long-term manual inspections are subject to visual fatigue and human errors. Even top manufacturers relying on large re-inspection teams face high labor costs and limited efficiency.

\begin{figure*}[!t]
	\centering
	\includegraphics[width=0.98\linewidth]{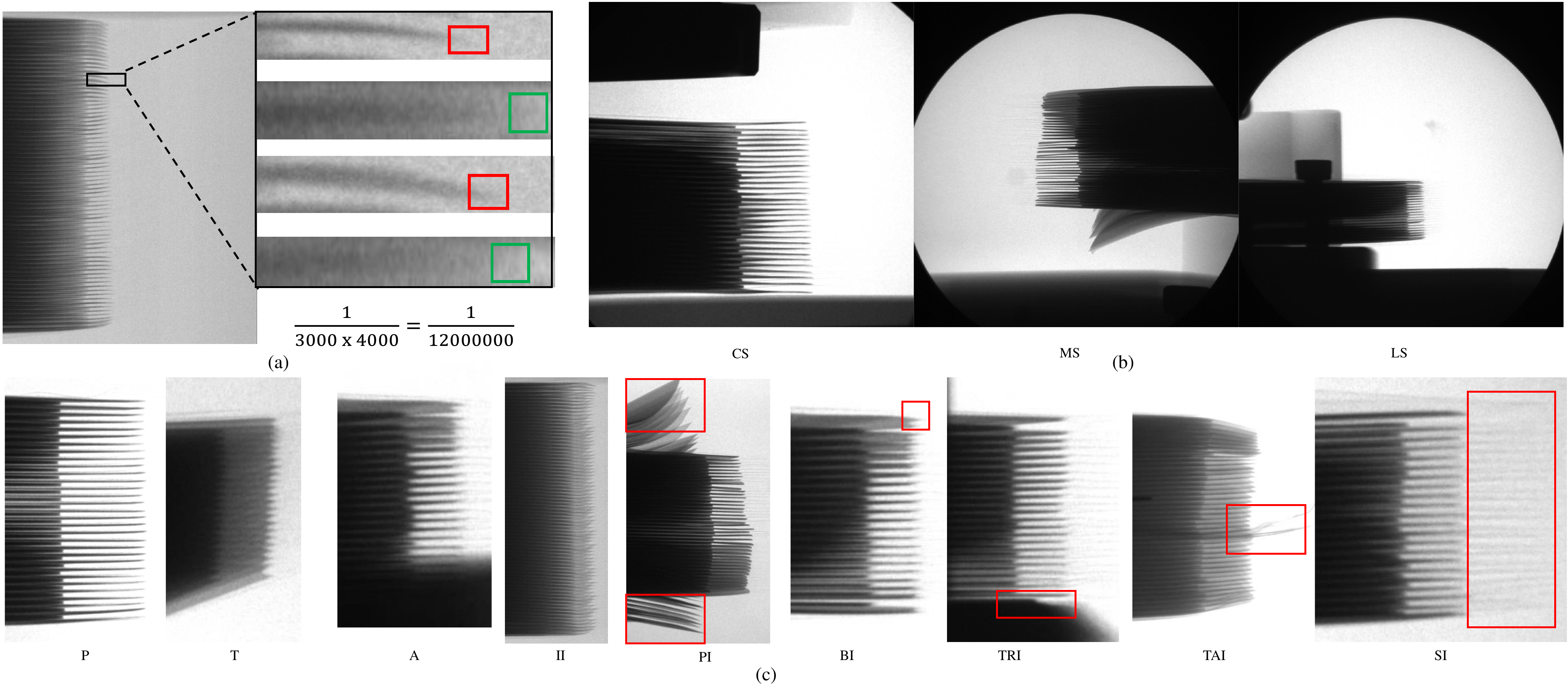}
	\caption{Visual examples from the PBD5K dataset (best viewed zoomed in). 
    (a) 1-pixel localization on high-resolution (3000 × 4000) X-ray images. 
    (b) The imaging field-of-view include Close Shot (\textit{CS}), Medium Shot (\textit{MS}) and  Long Shot  (\textit{LS}).
    (c) Examples of various attributes. See Tab.~\ref{tab:attribute_description} for details.
    }
	\label{fig:buttary_class}
    \vspace{-3mm}
\end{figure*}

\begin{table}[!t]
\centering
 \caption{Attribute descriptions (see examples in Fig.~\ref{fig:buttary_class}(c)).}
\label{tab:attribute_description}
\resizebox{\linewidth}{!}{
    \setlength\tabcolsep{1pt}
    \renewcommand\arraystretch{1.2}
\begin{tabu}{c||ccccl}
	\toprule[2pt]
	Attr & &&&&Description\\
	\hline
	\textbf{P} &&&&& \textit{Pure Plate.} High-quality  sample without any internal and external interference.\\
	\textbf{T} &&&&& \textit{Tilted Plate.} Deformation caused by excessively dense plates.\\
	\textbf{A} &&&&& \textit{Aberrant Plate.} 1) Plates visualization is incomplete due to occlusion.\\
	&&&&& 2) Plates are not arranged in the order of anode $\rightarrow$ cathode $\rightarrow$ anode. \\
	\textbf{II} &&&&& \textit{Illumination  Interference.} Visual thickness anomaly of the plate caused by \\
	&&&&& deviations in the incident angle of the radiation source.\\
	\textbf{PI} &&&&& \textit{Plate Interference.} Other batteries blend into current view.\\
	\textbf{BI} &&&&& \textit{Bifurcation Interference.} A single plate produces a bifurcation.\\
	\textbf{TRI} &&&&& \textit{Tray Interference.} A battery tray is a device used for holding large battery packs in place. \\
	&&&&& Close distance between the side plate and tray may cause visual interference.\\
	\textbf{TAI} &&&&& \textit{Tab  Interference.} Battery tabs are the anode and cathode connectors that \\
	&&&&& carry the cells' electrical current.  \\
	\textbf{SI} &&&&& \textit{Separator Interference.} A battery separator is a type of polymeric membrane \\
	&&&&& that is positioned between the anode and cathode. \\
	
	\bottomrule[2pt]
\end{tabu}
	}
        \vspace{-3mm}
\end{table}

Therefore, the development of an intelligent PBD model is an imminent need. 
To benchmark this new task, we first introduce PBD5K, the first large-scale power battery dataset with 5,000 X-ray images. It is constructed using an intelligent annotation pipeline that combines automatic filtering, model-assisted labeling, cross-validation, and hierarchical quality control, significantly improving labeling efficiency and accuracy. 
Compared with traditional visual tasks, PBD presents several unique challenges: 
\textit{\textbf{{\uppercase\expandafter{\romannumeral1})}}}  \textit{\textbf{Open Vision Problem Modeling.}} 
Like emerging tasks such as character stroke extraction~\cite{CCSE}, insubstantial object detection~\cite{IOD}, open vocabulary object detection and segmentation~\cite{OVD,OVSeg}, how to model a suitable AI-based solution is critical for the PBD task. 
\textit{\textbf{{\uppercase\expandafter{\romannumeral2})}}}   \textit{\textbf{Single Pixel-Level Object Localization.}}
As shown in Fig.~\ref{fig:buttary_class}(a), PBD requires accurately identifying plate endpoints that are only one pixel in size among millions of pixels, and providing their precise coordinates. This level of search difficulty is akin to finding a needle in a haystack.  
\textit{\textbf{{\uppercase\expandafter{\romannumeral3})}}}  \textit{\textbf{Multi-shot Imaging.}} 
As shown in Fig.~\ref{fig:buttary_class}(b), X-ray imaging is conducted under various field-of-view settings, leading to significant variations in spatial resolution. Even under the same imaging setup, batteries may exhibit different physical dimensions or placement scales. 
\textit{\textbf{{\uppercase\expandafter{\romannumeral4})}}}  \textit{\textbf{Weak Feature Perception.}} 
As clearly observed in Fig.~\ref{fig:buttary_class}(c), the plate endpoints exhibit extremely weak features and are visually similar to their surrounding regions. Especially under interference from bifurcated plates, separators, or trays, their visibility is further diminished. 
\textit{\textbf{{\uppercase\expandafter{\romannumeral5})}}}    \textit{\textbf{Fine-grained Structural Semantics.}} 
PBD5K contains a wide range of overhang distribution patterns. When the distance between adjacent cathode and anode plates is small (less than 100 pixels), their visual appearances become easily confused. 
Since power batteries are manufactured with strict alternation between cathode and anode plates, accurate endpoint localization implicitly requires the model to understand the structural semantics, including the classification and ordering of electrode plates.

 \begin{figure}[t]
	\centering
	\includegraphics[width=0.96\linewidth]{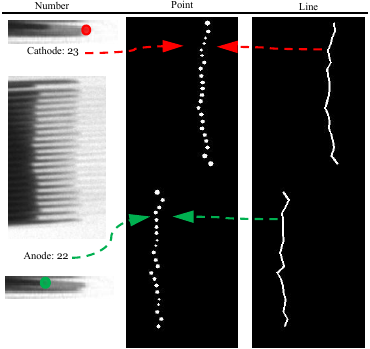}
	\caption{Illustration of multi-clue mining with  point, line and number clues.
}
	\label{fig:multicues}
        \vspace{-3mm}
\end{figure}

In this paper, we propose a multi-dimensional collaborative network for accurate PBD. Imitating the coarse-to-fine human visual perception system~\cite{coarse-to-fine1,coarse-to-fine2}, the model first perceives the battery region and then gradually refines the localization of each plate. Once the endpoints are precisely segmented, both their coordinates and total numbers can be derived.
\textbf{\textit{First}}, we reformulate PBD as a point-centric segmentation task, where the core objective is to predict a point map of single-pixel endpoints for cathode and anode plates. A U-shaped encoder-decoder~\cite{FPN,Unet} serves as the basic structure for point prediction. Inspired by multi-task learning~\cite{Hard_parameter_sharing,soft_parameter_sharing,MMFT} and the structural traits of power batteries, we introduce two auxiliary tasks: line segmentation and plate counting. 
As shown in Fig.~\ref{fig:multicues}, connecting predicted endpoints yields line maps that constrain spatial position and reduce ambiguity. Line supervision guides endpoint refinement, while the counting branch enforces consistency between the predicted and actual number of electrodes, mitigating false positives and omissions. By jointly learning from these three complementary clues, we establish a segmentation-centered multi-dimensional collaborative framework. 
\textbf{\textit{Next}}, to improve localization under weak textures and noisy backgrounds, we utilize state space models (SSMs)~\cite{LSSL,S4,Mamba,VMamba}, which offer efficient long-range modeling with linear complexity. Compared to Transformer-based models~\cite{ViT,Swin}, SSMs are more suitable for high-resolution industrial images due to their lower computational cost. 
However, direct use of SSMs may amplify irrelevant large-region patterns (\textit{e.g.}, trays or tabs), thereby overwhelming the subtle and sparse endpoint signals.
To address this, we propose the prompt-filtered state space module (PFSSM). It uses features from a representative prompt image to generate dynamic filters that enhance plate features while suppressing distractors. Filtered features are then processed through a 2D Mamba block~\cite{VMamba} for context modeling. This design improves robustness under low contrast, occlusion, or interference, directly addressing the challenges of \textit{weak feature perception} and \textit{fine-grained structural semantics}. 
\textbf{\textit{Third}}, we embed the density-aware reordering state space module (DRSSM) at the tail of MDCNeXt to refine predictions in densely packed regions. General sequential scanning in SSM often mixes background and foreground features, fragmenting the already sparse electrode signals. This problem worsens in samples with densely packed plates, where spatial entanglement further weakens class-specific representations. DRSSM addresses this by using the coarse point map to semantically regroup features (\textit{e.g.}, cathode, anode, background) and reorder them into contiguous sequences. These reordered tokens are processed via state space modeling to enhance intra-class consistency. The refined features are then mapped back to their original layout, yielding sharper boundary localization and better category separation.
\textbf{\textit{Last}}, we design an adaptive label generation strategy for point segmentation. Specifically, we compute the spatial distance between adjacent homopolar plates based on annotations and use it as a reference radius to generate scale-aware ground truth masks. 
Unlike fixed-size masks, this adaptive mask better matches real spacing variations, improving robustness in both sparse and dense plate regions.

Our main contributions can be summarized as follows: 
\begin{itemize}[leftmargin=*,itemsep=0em,topsep=0em,parsep=0em]

\item We introduce power battery detection (PBD) as a new visual task in computer vision and construct the first benchmark dataset PBD5K, design an effective baseline, formulate comprehensive metrics, and explore label generation strategies to promote research on the PBD.

\item We develop an efficient annotation pipeline for PBD5K that integrates automatic filtering, model-assisted pre-labeling, cross-validation, and multi-level quality control. 

\item We formulate PBD as a segmentation problem and propose MDCNeXt, a multi-dimensional collaborative framework that jointly leverages point, line, and number clues to improve both low-level details and high-level semantics.

\item We successfully adapt state space modeling to MDCNeXt by introducing prompt filtering and density-aware reordering mechanisms, enabling its robust localization under weak representation and dense regions.

\item We compare MDCNeXt with  corner detection, crowd counting, general/tiny object detection and image segmentation methods. Extensive experiments demonstrate that the proposed MDCNeXt performs favorably against the state-of-the-art methods under different metrics. 

 \item We provide systematic discussion and forward-looking outlook on some promising future directions for the PBD task, laying the foundation for long-term research and real-world deployment of automated battery quality inspection.
\end{itemize}

\textit{This paper is based on and extends our CVPR version~\cite{MDCNet} in the following aspects. 
\textbf{\uppercase\expandafter{\romannumeral1})}  We significantly expand the dataset from 1,500 to 5,000 images, forming a new benchmark PBD5K with richer variations in image resolution, plate number, overhang distribution and interference types, and provide a more detailed dataset analysis.
\textbf{\uppercase\expandafter{\romannumeral2})}  An intelligent annotation pipeline is introduced to improve labeling efficiency and consistency, providing the industrial AI community with a transparent and reproducible workflow. 
\textbf{\uppercase\expandafter{\romannumeral3})} We upgrade the MDCNet~\cite{MDCNet} to  a stronger version named MDCNeXt by embedding the proposed prompt-filtered state space module and density-aware reordering state space module.
\textbf{\uppercase\expandafter{\romannumeral4})} We conduct more extensive experiments, including quantitatively and qualitatively comparisons with generalist, general and specialized segmentation models in terms of ten metrics, showing consistent superiority of MDCNeXt.
\textbf{\uppercase\expandafter{\romannumeral5})} We provide a comprehensive introduction to related work,  detailed implementation details, and thorough ablation studies. 
\textbf{\uppercase\expandafter{\romannumeral6})} 
We present an in-depth discussion on future directions of PBD, outlining promising research opportunities.
}

\section{Related Work}

\subsection{Open Vision Problem Modeling}

Emerging open vision problems often deviate from traditional tasks due to novel instance definitions, annotation challenges, or perceptual ambiguity, demanding tailored task abstraction and modeling. 
Liu \textit{et al.}~\cite{CCSE} formulate \textbf{Chinese character stroke extraction} as \uline{stroke-level instance segmentation}, enabling fine-grained and transferable structural decomposition.
Zhou \textit{et al.}~\cite{IOD,GOD} address the challenge of \textbf{insubstantial object detection} by treating faint or transparent targets as \uline{spatially diffuse regions with geometric uncertainty}, and model detection through 3D voxel shift field estimation.
Pang \textit{et al.}~\cite{OVCOS} define \textbf{open-vocabulary camouflaged object segmentation} as a hybrid of \uline{open-set recognition} and \uline{camouflage-aware segmentation}, integrating vision-language priors and structural clues like depth and edges to capture imperceptible objects.
Chiu \textit{et al.}~\cite{WireSegHR} frame \textbf{wire segmentation} as a \uline{thin-structure semantic segmentation} task under ultra-high-resolution constraints, emphasizing continuity-preserving modeling tailored for elongated, slender targets.
Yang \textit{et al.}~\cite{TSI} conceptualize \textbf{traffic sign interpretation} as a \uline{semantic reasoning task over interrelated signs}, where the goal is to interpret visual signals into natural-language instructions through multi-level relational logic.
Zhou \textit{et al.}~\cite{NighttimeOpticalFlow} tackle the \textbf{nighttime optical flow estimation} problem by casting it as \uline{cross-domain motion} alignment, leveraging reflectance and gradient-aligned latent spaces to transfer motion priors from auxiliary domains under degraded visibility.
Liu \textit{et al.}~\cite{VectorGraphNET} design \textbf{technical drawing parsing} as a \uline{vector-based structural graph learning} problem, transforming CAD designs into spatial graphs where geometric primitives are modeled as relational nodes for precise topological inference.
Wei \textit{et al.}~\cite{VecFormer} further leverage a \uline{type-agnostic and expressive line-based representation} of graphical primitives, instead of traditional point-based methods, enabling a unified abstraction that accommodates diverse elements in engineering and architectural diagrams.

These pioneering works show a common principle: solving underexplored visual problems starts with proper \textit{task redefinition} and \textit{representation design}, enabling meaningful perception and reasoning. 
Our work follows this line by introducing power battery detection as a new vision task, with a tailored representation and modeling strategy bridging real-world industrial needs and AI perception.

\subsection{Object Detection, Counting and Segmentation}
We introduce multiple potential modeling schemes for the PBD task and analyze related techniques. 

\textbf{Corner detection} is widely used in motion detection, image matching, and video tracking. A corner point is typically defined as the intersection of two edges and can also be referred to as a feature point. Harris and Stephens~\cite{Harris} propose one of the earliest methods by directly computing the differential of the corner score with respect to direction. Shi–Tomasi~\cite{Shitomasi} improves stability by selecting corners based on the minimum eigenvalue of the gradient matrix. Sub-pixel methods~\cite{Sub-Pixel} further enhance accuracy by refining corner positions to real coordinates for geometric measurement or calibration. However, corner detectors are often not robust and typically require high redundancy to mitigate the impact of individual errors on recognition tasks.

\textbf{Object detection} aims to locate and classify semantic objects in images or videos. Many methods rely on \textbf{bounding box prediction}, using rectangles to approximate object locations. Detectors are typically categorized into two paradigms: 
\textit{{\uppercase\expandafter{\romannumeral1})}} \emph{Two-stage detectors}, such as Fast RCNN~\cite{Fast_rcnn}, Faster R-CNN~\cite{Faster_rcnn}, FPN~\cite{FPN}, and RFCN~\cite{RFCN}, which generate region proposals before classification and refinement. 
\textit{{\uppercase\expandafter{\romannumeral2})}}  \emph{One-stage detectors}, such as RetinaNet~\cite{RetinaNet}, YOLOv10~\cite{Yolov10}, FCOS~\cite{FCOS}, and DETR~\cite{DETR}, which directly predict bounding boxes and categories in a single forward pass. 
Although most single-stage methods have fast detection speed, they are usually not as accurate as the anchor-based two-stage detectors, especially in small objects. To bridge this gap, a number of \textbf{tiny object detection} methods have been developed to achieve a trade-off between accuracy and efficiency, including C3Det~\cite{C3Det}, CFINet~\cite{CFINet}, and HS-FPN~\cite{HS-FPN}.

\textbf{Dense object counting} task is to infer the number of objects in dense scenarios. Most CNNs-based methods~\cite{DM,CUT_cc,scale_variation_cc1,density_map_cc1} predict a density map from a crowd image, where the summation of the density map is the crowd count. 
For the structures, some works are proposed to address scale variation~\cite{scale_variation_cc1,scale_variation_cc2,scale_variation_cc3}, to refine the predicted density map~\cite{density_map_cc1,density_map_cc2,IOCFormer}, and to encode context information~\cite{context_information_cc1,context_information_cc2}.
For the loss function, L2 loss is usually used in many models but it is sensitive to the choice of variance in the Gaussian kernel~\cite{Adaptive_CC}. Therefore, Bayesian loss~\cite{BL} is proposed with point-wise supervision.

\textbf{Image segmentation} aims to simplify or transform an image into a more structured representation, facilitating further analysis. 
Diverse segmentation types exist in practice, including salient, camouflaged, transparent, and medical segmentation. 
Most methods~\cite{DeepLabV3+,MINet,MSNet_Polyp} adopt a \uline{U-shaped encoder-decoder architecture}~\cite{Unet,FPN}, where multi-level features are progressively fused to generate high-resolution masks. 
To enhance feature representation, some works~\cite{AMP,BDRAR_Shadow,GateNet,li2023delving} incorporate \uline{attention mechanisms}, such as spatial, channel-wise  and gated attention to selectively emphasize informative clues. Meanwhile, handling \uline{scale variation} among objects remains a key challenge. This has motivated research into multi-scale feature extraction techniques~\cite{GateNetv2,ASPP,DenseASPP,M2SNet,ZoomNet}, including ASPP~\cite{ASPP}, Fold-ASPP~\cite{GateNet}, and DenseASPP~\cite{DenseASPP}, which aim to capture semantic clues across receptive fields of varying sizes. 
Recently, \uline{Transformer-based models}~\cite{Segmenter,SegFormer,SwinUNet} have attracted attention for their ability to model global context. Unlike CNNs with limited receptive fields, Transformers~\cite{transformer} utilize self-attention to capture long-range dependencies, improving performance in complex spatial tasks.  Segformer~\cite{SegFormer} and SwinUNet~\cite{SwinUNet} couple hierarchical encoding with attention-based decoding. However, their quadratic complexity and weak local bias limit scalability and fine-grained prediction. 
In contrast, \uline{Mamba-based methods}~\cite{Sigma,Segmamba} leverage selective state-space models to replace attention with structured recurrence, enabling linear time complexity and superior scalability. Mamba~\cite{Mamba} excels in global dependency modeling with low memory cost and an inherent structural bias, suiting thin-structure or topology-aware tasks.  Some hybrid \uline{CNN-Mamba} models~\cite{U-mamba,CM-UNet} combine local detail preservation with global reasoning, and providing an efficient framework for high-resolution and real-time segmentation.
With the growing pursuit of foundation models and artificial general intelligence, \uline{unified and generalist segmentation models} are emerging to address diverse segmentation tasks with a single set of parameters.
SAM~\cite{SAM} is trained on 1.1 billion masks, enabling high-quality segmentation with various prompts.
UniverSeg~\cite{UniverSeg} focuses on unifying medical image segmentation across diverse organs and modalities.
SegGPT~\cite{SegGPT} offers flexible segmentation via visual in-context learning.
HQSAM~\cite{HQSAM} improves SAM for high-resolution segmentation.
SAM 2~\cite{SAM2} extends SAM to video by incorporating memory attention and multi-frame prompts.
For complex context-dependent concepts, EVP~\cite{EVP} and GateNetv2~\cite{GateNetv2} integrate low-level structural prompts and gated context, while Spider~\cite{CDCU-Spider} and VSCode~\cite{VSCode} apply 2D \uline{prompt learning} to model foreground-background relations.
In this work, our proposed MDCNeXt integrates the aforementioned advanced techniques, including CNN-Mamba hybrid design, attention-based refinement, multi-scale modeling, and prompt-driven filtering, aiming to provide a high-precision segmentation for the PBD task.

\subsection{Enhancing Weak Features in Complex Vision Tasks}
In complex vision tasks, key targets often exhibit weak, sparse, or ambiguous features due to small object scales, low contrast, occlusion, vague boundaries, or background clutter. Such weak signals pose fundamental challenges in remote sensing, medical imaging and industrial inspection. 
To address this, recent studies enhance weak features via two main strategies: multi-modal and multi-structural clues fusion, both injecting complementary or contextual information into feature learning. 
Multi-modal fusion improves feature expressiveness by introducing multiple modalities. For example, Liu \textit{et al.}~\cite{RGBDCOD} employ RGB-D for camouflaged object segmentation, Kim \textit{et al.}~\cite{Transpose} utilize RGB-T for transparent object segmentation, and Pang \textit{et al.}~\cite{CAVER_RGBDSOD} verify the complementarity between RGB-D and RGB-T modalities. 
Beyond visual signals, text serves as a strong auxiliary modality in semantically ambiguous scenes. CLIP~\cite{CLIP} enable open-vocabulary recognition via large-scale image-text pretraining, while Pang \textit{et al.}~\cite{OVCOS} use textual descriptions to generalize camouflaged object segmentation to novel categories. In the medical domain, cross-modality fusion (\textit{e.g.}, MRI, CT, PET) provides complementary anatomical and functional context for localizing subtle lesions~\cite{Multi-modality_fusion_survey}. 
Beyond modalities, structural priors such as edges, contours, and symmetry help delineate fuzzy or ambiguous targets. Edge-aware modules~\cite{Inf-Net,CFANet_Polyp,EGNet} guide fine-grained segmentation, while higher-level topological clues (\textit{e.g.}, skeletons, geometric primitives) benefit tasks like polyp segmentation~\cite{GSNet_Polyp}, shadow removal~\cite{DAS_shadow_removal}, and transparent object detection~\cite{Trans2Seg_Transparent}. These structures isolate meaningful patterns amid visual noise.
Our work aligns with this direction by constructing a multi-clue segmentation framework that jointly leverages point, line, and number clues to enhance weak feature perception, enabling robust object discovery in X-ray battery images with complex interference.

\section{PBD5K Dataset}
\subsection{Image Collection and Annotation}
\begin{figure*}[!t]
	\centering
	\includegraphics[width=\linewidth]{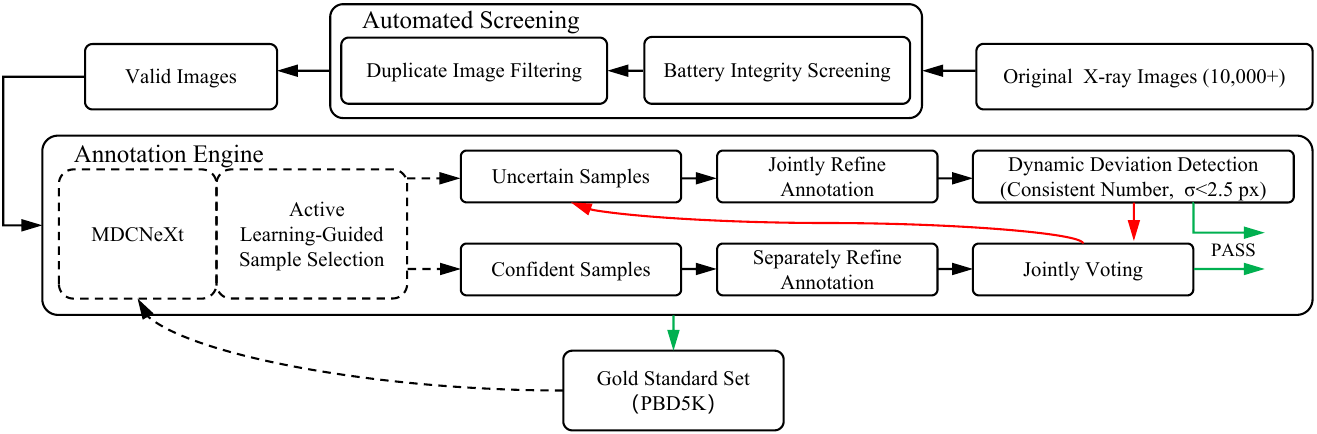}
	\caption{The overall intelligent annotation pipeline with automated screening, annotation engine with active Learning, and multi-expert  quality control. \textcolor{red}{Red arrows} (\textcolor{red}{$\longrightarrow$})  indicate poor annotation quality, while \textcolor{mygreen}{green arrows} (\textcolor{mygreen}{$\longrightarrow$})  
    indicate high-quality annotations that are included in the Gold Standard Set. }
	\label{fig:automatic_annotation}
\end{figure*}

The proposed PBD5K dataset is collected using the same DR device across over 10,000 power battery cells from $10$ manufacturers, covering close, medium, and long shot views.
To efficiently construct a high-quality industrial dataset, we design an intelligent annotation pipeline that integrates automated screening, uncertainty-guided active learning, and multi-expert quality control.
The full workflow is illustrated in Fig.~\ref{fig:automatic_annotation}, where over 10,000  raw X-ray images are progressively filtered, annotated, and validated to form the final PBD5K benchmark with 5,000 images.

\textbf{{Automated Screening.}} 
The pipeline begins with the automated filtering of raw data. 
This stage includes battery integrity screening and duplicate image filtering.
The former employs a lightweight object detection model to discard invalid samples with incomplete views (\textit{e.g.}, batteries cut off at edges), overexposed images, and blank captures.
The latter eliminates near-duplicate images by computing high-level visual similarity using pre-trained convolutional neural network (CNN) features. 
Specifically, each image is first encoded into a compact 512-dimensional representation using global average pooling over intermediate VGG~\cite{VGG} feature maps.
To handle large-scale retrieval, we apply FAISS~\cite{FAISS}, a fast approximate nearest neighbor search library. Image pairs with Euclidean distance below a threshold are deemed duplicates. We then perform connected-component clustering on the similarity graph and retain one representative per cluster. 
This multi-stage screening ensures the final dataset is both structurally complete and semantically diverse.

\textbf{{Annotation Engine with Active Learning.}}  
{Model-assisted annotation is a widely accepted practice in large-scale dataset construction. Well-known foundation models such as SAM~\cite{SAM}, SAM 2~\cite{SAM2}, and SAM 3~\cite{SAM3} adopt model-in-the-loop pipelines that combine initial manual annotation, model-generated proposals, human refinement, and voting or filtering mechanisms.} 
In this work, the annotation engine is driven by our MDCNeXt within an active learning loop. 
Initially, with no labeled samples, all valid images are sent to a multi-expert annotation stage. Each sample is independently labeled by three experts. A real-time deviation detection module checks consistency in point counts and spatial locations.
If the annotations are consistent in number and have low position deviation, the system fuses them via coordinate averaging to form a gold standard. Otherwise, the sample proceeds to a voting stage with three additional experts to select the most plausible annotation. To avoid infinite review loops, we place constraints on voting depth. 
Once enough gold-standard labels are collected, MDCNeXt is trained and used to predict on unlabeled images.
 To assess the uncertainty of the coarse predictions, we evaluate the degree of binarization of the predicted maps. Specifically, we compute the difference between the maximum and minimum activation values of the sigmoid-activated map (ranging from 0 to 1). 
Based on this uncertainty estimation, the predicted samples are divided into two categories:
\begin{itemize}[leftmargin=*,itemsep=0em,topsep=0em,parsep=0em]
    \item \textbf{Uncertain Samples:} Predictions with a larger difference are routed back to the joint annotation pipeline, where multiple experts collaboratively review and revise the results.
    \item \textbf{Confident Samples:} Predictions with a lower difference are refined independently by individual experts. These refined annotations are then passed through the jointly voting stage for final validation and integration.
\end{itemize}

\begin{table}[!t]
	\centering
	\caption{Efficiency comparison between manual and intelligent annotation schemes.}
	\label{tab:annotation_efficiency}
	\resizebox{\linewidth}{!}{
		\setlength\tabcolsep{1pt}
		\renewcommand\arraystretch{1}
		\begin{tabular}{l||ccl}
			\toprule[2pt]
			{Metric} & {Manual} & {Intelligent} & {Improvement} \\
			\hline
			Average Annotation Time (min/image) & 3.4 & 1.2 & $\uparrow$64.7\% \\
			Annotation Spatial Deviation (pixel)  & 3.5 & 0.9 & $\uparrow$74.3\% \\
			Proportion of Disputed Samples & 22\% & 13\% & $\uparrow$40.9\% \\
			Rework Rate & 17\% & 6\% & $\uparrow$64.7\% \\
			\bottomrule[2pt]
		\end{tabular}
	}
\vspace{-1em}
\end{table}

During the entire process, human expert teams work in parallel rather than sequentially, ensuring high throughput. This active learning-driven annotation loop iteratively improves MDCNeXt and continues until convergence.
{All model-assisted proposals undergo strict human expert verification at the point level. Annotators do not simply accept model predictions. Instead, each plate endpoint is carefully reviewed, with experts performing zoom-in inspection, cross-comparison across frames, and alignment with physical bare-cell references when necessary. Importantly, the annotation standard remains consistent throughout the entire dataset construction process. Before model assistance is introduced, annotations are fully manual, and after introducing MDCNeXt, the same quality criteria and precision requirements are strictly maintained.} 
Through this pipeline, we construct the {PBD5K} dataset, a gold-standard benchmark composed of 5,000 expertly annotated X-ray battery images, validated  through automated and human-in-the-loop quality checks. As shown in Tab.~\ref{tab:annotation_efficiency}, our approach significantly improves annotation efficiency and reliability compared to traditional manual schemes.

 \begin{figure}[!t]
	\centering
	\includegraphics[width=\linewidth]{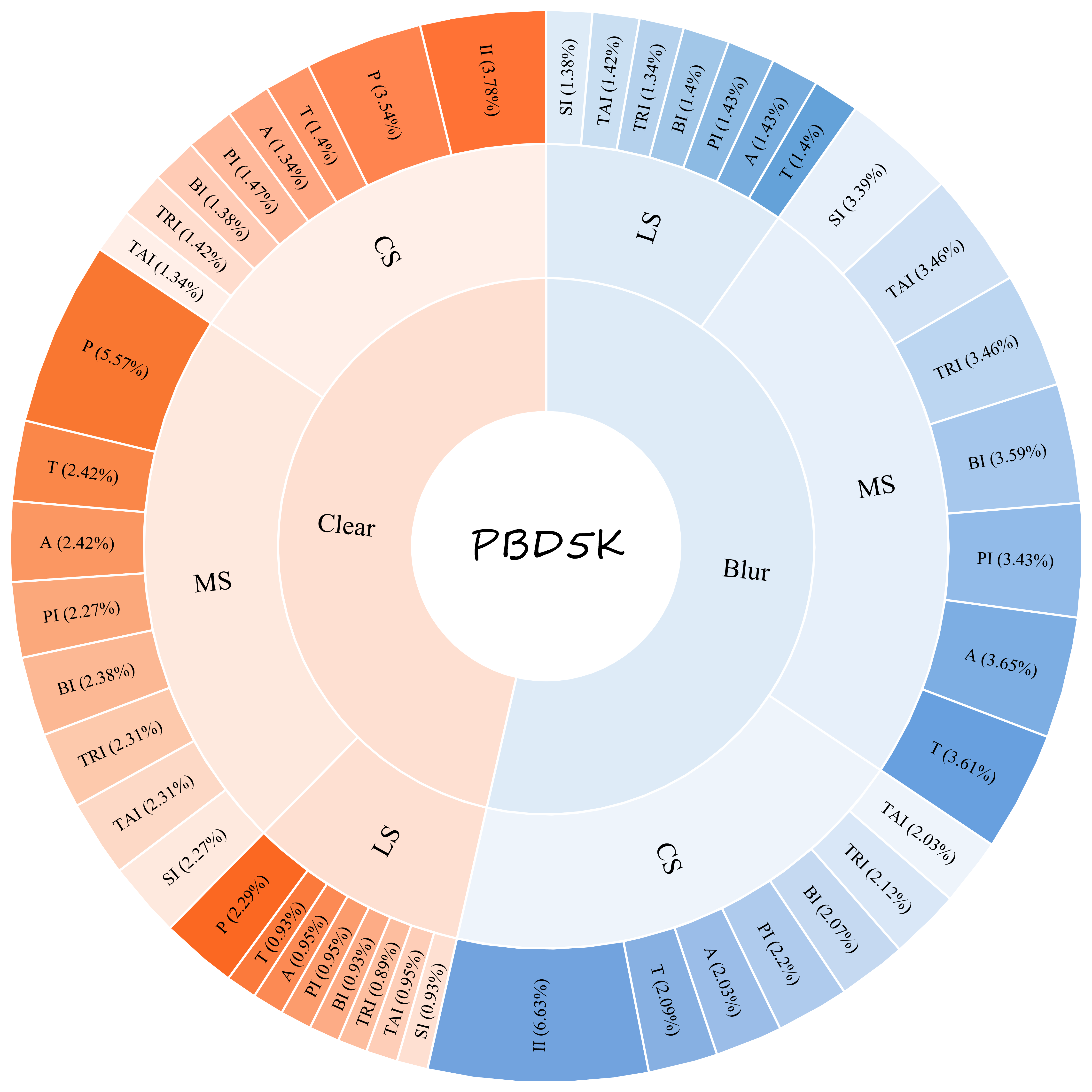}
	\caption{Hierarchical distribution of categories, shots, and attribute types in the PBD5K dataset.}
	\label{fig:PBD5K}
\end{figure}

\subsection{Dataset Analysis}
\textbf{Hierarchical Distribution of Categories and Attributes.}
As shown in Fig.~\ref{fig:PBD5K}, the collected images have 2 categories, 3 shots, 9 attributes and their proportions under each shot type are statistically reported in detail. Different battery types, manufacturing and packaging processes produce
different thicknesses of batteries. Even if the DR device
is calibrated, it still has different degrees of penetration for different thicknesses, thereby resulting in \textit{Clear} and \textit{Blur} X-ray images. The imaging field-of-view include  close shot (\textit{CS}), medium shot (\textit{MS}) and  long shot  ({\textit{LS}). Each shot type further contains samples labeled with multiple common attributes: pure plate (\textbf{P}), tilted plate  (\textbf{T}), aberrant plate (\textbf{A}), illumination interference (\textbf{II}), plate interference  (\textbf{PI}), bifurcation interference (\textbf{BI}), tray interference  (\textbf{TRI}), tab interference (\textbf{TAI}), and separator interference  (\textbf{SI}).  The quantitative proportions are marked beside each attribute. We can see that \textbf{P} only appears in the \textit{Clear} category, \textbf{II} is exclusively found in the \textit{CS} view, while \textbf{SI} does not occur in the \textit{CS} view. The detailed definitions of these attributes are given in Tab.~\ref{tab:attribute_description}, and visual samples are presented in Fig.~\ref{fig:buttary_class}.

\textbf{Co-Occurrence Distribution of Attributes.}
Fig.~\ref{fig:PBD5K_heatmap} (left) shows a dependency matrix where each value indicates the conditional probability of co-occurrence. The highly non-diagonal and asymmetric structure reveals complex inter-attribute relationships. \textbf{SI} frequently coexists with many other attributes, acting as a common disturbance. In contrast, \textbf{TRI} and \textbf{TAI} rarely co-occur with others, indicating they are more isolated and category-specific attributes.  Some combinations, such as \textbf{BI + PI} or \textbf{T + A}, occur frequently due to their correlation in battery production anomalies.

\textbf{Multi-Dependency Relationships.}
 In Fig.~\ref{fig:PBD5K_heatmap} (right), we provide a chord diagram to further visualize mutual dependencies. The arc thickness represents the degree of interaction between attributes.  We can see that more than four attribute dependencies appear in all samples, which illustrates the diversity and complexity of this dataset.

 \begin{figure}[!t]
	\centering
	\includegraphics[width=\linewidth]{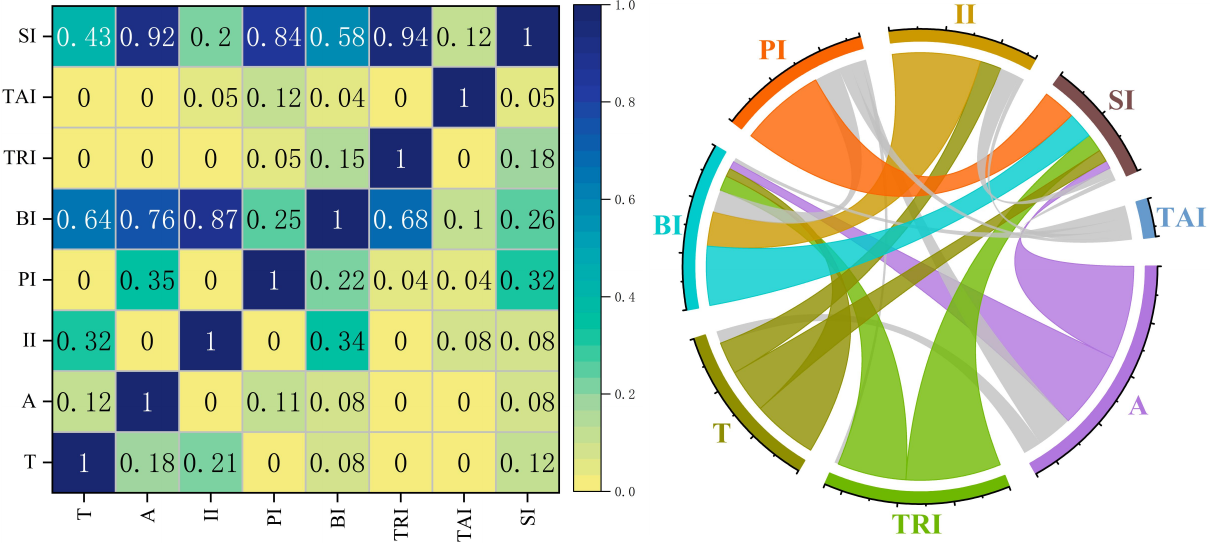}
	\caption{Left: Co-occurrence distribution of attributes. The numbers in each grid indicate the proportion of images. Right: Multiple dependencies among these attributes. The larger the arc length, the higher the probability that one attribute is related to another.}
	\label{fig:PBD5K_heatmap}
\end{figure}

 \begin{figure}[!t]
	\centering
	\includegraphics[width=\linewidth]{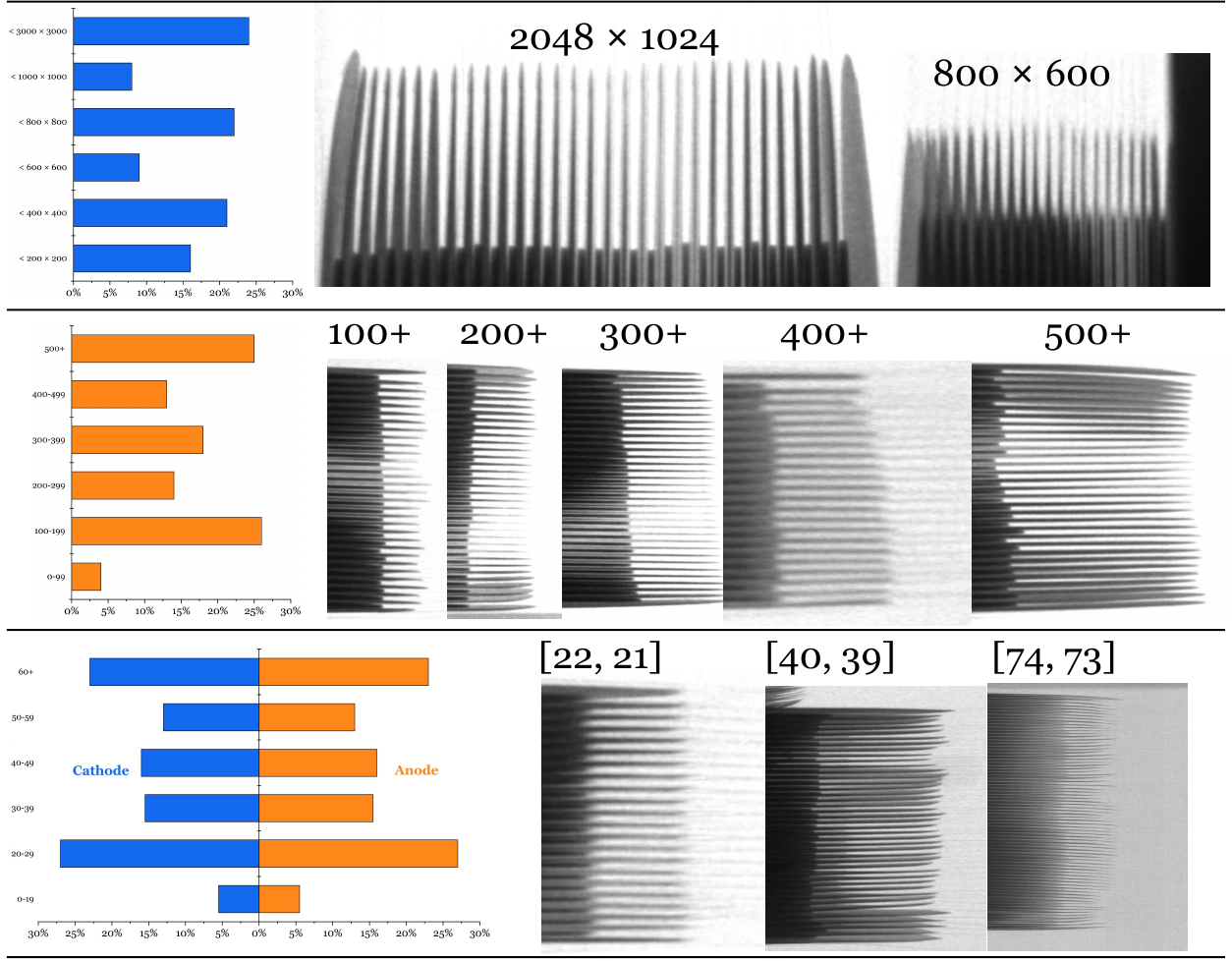}
	\caption{ Statistics and visualization of structural diversity in the PBD5K dataset, including image resolution, overhang and plate number distributions.}
	\label{fig:PBD_task_number_overhang_resolution}
\end{figure}

\textbf{Resolution, Overhang Distribution, and Plate Number.}
Fig.~\ref{fig:PBD_task_number_overhang_resolution} presents three key structural features. (1)  The dataset contains images of diverse resolutions, ranging from less than $200\times200$ to over $3000\times3000$ pixels. Representative examples (\textit{e.g.}, $2048\times1024$ and $800\times600$) are visualized alongside the histogram. (2) We provide a statistical distribution of overhang values. Five examples with different overhang levels, demonstrating the wide range of structural inconsistencies. This variability is essential for evaluating model robustness to structural deviations. (3) 
The number of cathode/anode plates in each battery can be fewer than 20 or exceed 60. Three examples with different plate numbers clearly illustrate that increasing the number of plates poses significant challenges for visual plate disentanglement.

\textbf{Dataset Splits.}
To ensure balanced coverage of various characteristics, we split the dataset into training and testing sets with a 6:4 ratio while maintaining distribution balance across categories, shots, and attributes. There are 3,000 images for training and 2,000 for testing. Furthermore, we split the test set into three difficulty levels: regular (515 images), difficult (808 images), and tough (677 images), based on the degree of interference caused by different attributes and plate number.

\subsection{Evaluation Metrics}
According to the number and overhang criteria adopted by the manufacturers, we design eight complementary metrics to quantitatively evaluate the performance of algorithms. Specifically, they are number mean absolute error (AN-MAE, CN-MAE), number accuracy (AN-ACC, CN-ACC, PN-ACC) and position mean absolute error (AL-MAE, CL-MAE and OH-MAE) for the anode level, cathode level, and pair level, respectively. 
Let \( n_i^x \) and \( \hat{n}_i^x \) represent the predicted and ground truth plate numbers for the \( i \)-th image at level \(x \in \{\text{anode, cathode,}\\\text{pair}\}\), and \( p_{i,j}^y \), \( \hat{p}_{i,j}^y \) represent the coordinates of the \( j \)-th  plate at level \( y \in \{\text{anode, cathode}\} \) for image \( i \) with resolution \( H_i \times W_i \).
The eight evaluation metrics are defined as follows:

\newcommand{\deltaP}[2]{\frac{1}{H_{#1} W_{#1}} \left|#2\right|}

\begin{equation}
\text{AN-MAE} = \frac{1}{N} \sum_{i=1}^{N} \left| n_i^{an} - \hat{n}_i^{an} \right|,
\end{equation}

\begin{equation}
\text{CN-MAE}  = \frac{1}{N} \sum_{i=1}^{N} \left| n_i^{ca} - \hat{n}_i^{ca} \right|,
\end{equation}

\begin{equation}
\text{AN-ACC}  = \frac{1}{N} \sum_{i=1}^{N} \mathds{1}\left(n_i^{an} = \hat{n}_i^{an} \right),
\end{equation}

\begin{equation}
\text{CN-ACC}  = \frac{1}{N} \sum_{i=1}^{N} \mathds{1}\left(n_i^{ca} = \hat{n}_i^{ca} \right),
\end{equation}

\begin{equation}
\text{PN-ACC}  = \frac{1}{N} \sum_{i=1}^{N} \mathds{1}\left(n_i^{pair} = \hat{n}_i^{pair} \right),
\end{equation}

\begin{equation}
\text{AL-MAE}  = \frac{1}{N_{p}} \sum_{i=1}^{N} \frac{1}{n_i^{an}} \sum_{j=1}^{n_i^{an}} \deltaP{i}{p_{i,j}^{an} - \hat{p}_{i,j}^{an}},
\end{equation}

\begin{equation}
\text{CL-MAE}  = \frac{1}{N_{p}} \sum_{i=1}^{N} \frac{1}{n_i^{ca}} \sum_{j=1}^{n_i^{ca}} \deltaP{i}{p_{i,j}^{ca} - \hat{p}_{i,j}^{ca}},
\end{equation}

\begin{equation}
\text{OH-MAE}  = \frac{1}{N_{p}} \sum_{i=1}^{N} \frac{1}{n_i^{an}} \sum_{j=1}^{n_i^{an}} \deltaP{i}{o_{i,j} - \hat{o}_{i,j}},
\end{equation}

\begin{equation}
o_{i,j} = \left| p_{i,j}^{cathode} - p_{i,j}^{anode} \right| + \left| p_{i,j}^{cathode} - p_{i,j+1}^{anode} \right|,  
\end{equation}
where \( N \) is the total number of test samples and \( N_{p} \) is the number of samples with correctly predicted plate numbers. 
{
For the overhang value $o_{i,j}$, it is defined based on the relative alignment between one cathode plate and its two neighboring anode plates. The first term measures the offset between the $j$-th cathode and its directly corresponding anode, while the second term captures the offset to the adjacent anode on the other side.
This formulation reflects the physical definition of overhang in battery electrodes, where a cathode plate is sandwiched between two neighboring anode plates.}
Note that, for an image, only when its number of AN or CN is predicted with 100\% accuracy, we can further evaluate the corresponding position MAE. 
Coordinates must be sorted before calculating AL-MAE, CL-MAE and OH-MAE. 

Generally,  $n_i^{x}$ can be obtained by counting the number of  $p_{i, j}^{{y}}$. Corner detection directly predicts each endpoint's coordinates. General/Tiny object detection methods predict bounding boxes for endpoints, and we calculate the center coordinates of each box as $p_{i, j}^{{y}}$. 
Counting methods estimate $n_i^{x}$   by summing predicted density maps. Segmentation-based methods produce point masks, from which $p_{i, j}^{{y}}$ is obtained by calculating the center coordinates for the circumscribed rectangle of each point map.

 \begin{figure*}[!t]
	\centering
	\includegraphics[width=\linewidth]{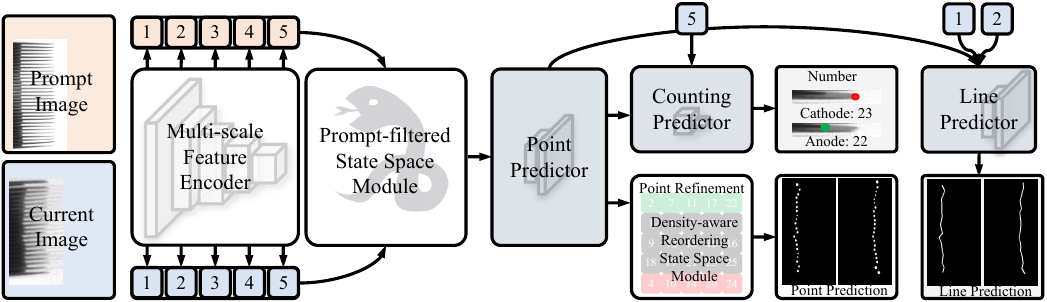}
	\caption{Overview of our MDCNeXt framework. It adopts an encoder-decoder architecture with a shared ResNet-50 backbone to extract five-level features from both the prompt and current images. The prompt-filtered state space module (PFSSM) utilizes prompt features to remove interference from the current features, generating clean plate representations. The point predictor produces coarse segmentation maps through five decoding layers. The line and counting predictors serve as auxiliary branches to enhance point prediction at low-level detail and semantic levels, respectively. The density-aware reordering state space module (DRSSM) refines both features and coarse predictions to generate the final fine-grained point map.} 
	\label{fig:MDCNeXt}
\end{figure*}

\section{Proposed Framework}

\subsection{Overview}
As shown in Fig.~\ref{fig:MDCNeXt}, the MDCNeXt adopts the encoder-decoder structure with a ResNet-50~\cite{Resnet} backbone. The encoder initially extracts five-level features, and {the prompt and current pipelines share the same encoder  weights that are jointly updated during training.}
The prompt-filtered state space module (PFSSM) uses the prompt features to filter out interfering information from the current features and generate pure plate appearance features. The point predictor outputs the coarse point segmentation map after passing through five decoder blocks. 
The line predictor and counting predictor assist in point prediction at the low-level detail and semantic levels, respectively. Finally, the density-aware reordering state space module (DRSSM) carefully refines the coarse point segmentation map and features to obtain a refined point map prediction.

\subsection{Prompt-filtered State Space Module} 
As shown in Fig.~\ref{fig:PFSM}, PFSSM consists of a prompt-based feature filter followed by a state space modeling block. It takes the feature map from the multi-scale feature encoder as input and outputs enhanced features for the point predictor. PFSSM aims to suppress irrelevant patterns (\textit{e.g.}, trays, tabs, and bifurcations) while capturing long-range dependencies essential for accurate plate localization. 
We randomly choose a \textbf{P} plate sample (see Fig.~\ref{fig:buttary_class}) as the prompt image. From the prompt’s 1$^{\text{st}}$–5$^{\text{th}}$ layer features, we extract multi-level information to construct dynamic filters. We take both the 5$^{th}$ prompt and current image features as an example to illustrate the process of prompt filtering.  
We first conduct global average pooling and convolution for $F_{prompt}$. Then, the softmax function distributes  $F_{prompt}$ with the channel-wise soft attention. Next, we initialize the  $3\times3$ convolution with trainable parameters and multiply the soft attention to generate the aggregated parameters as the weights. 
The current feature map $F_{\text{current}}$ is then filtered by these weights using \texttt{Bconv3-D1}. 
The filtered features $F_{\text{filtered}}$  are passed through  a \texttt{SiLU} activation and a  2D Selective Scan  (SS2D)~\cite{VMamba}.  SS2D performs to bridge 1D array scannitng and 2D plane traversal, enabling the extension of selective SSMs to process vision data.  This allows prompt information to be globally propagated throughout the current feature map in a content-aware and spatially aligned manner. 
Finally, the output is passed through layer normalization (\texttt{LN}) and a linear transformation to produce the final enhanced features. Overall, PFSSM serves as a prompt-driven enhancement module that both condenses high-level guidance from prompt features and establishes strong long-range dependencies between prompt and current images in an efficient manner.
 \begin{figure}[t]
	\centering
	\includegraphics[width=\linewidth]{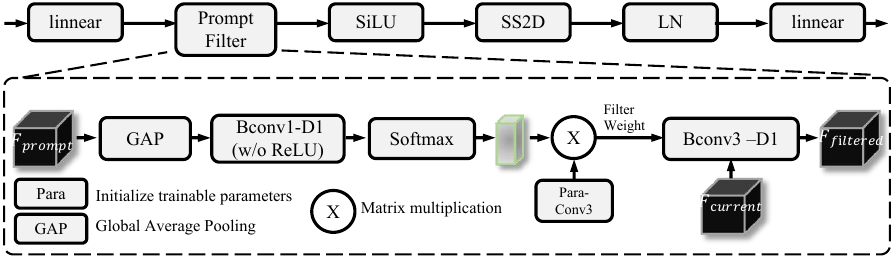}
	\caption{Illustration of the prompt-filtered state space module. It takes multi-level current and prompt features as inputs.}
	\label{fig:PFSM}
\end{figure}

 \begin{figure}[!t]
	\centering
	\includegraphics[width=\linewidth]{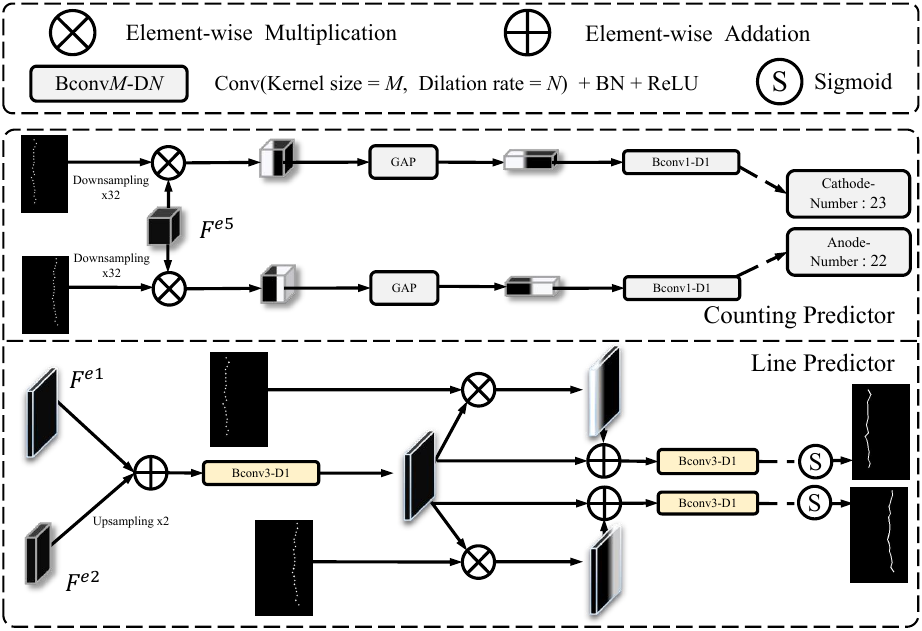  }
	\caption{Illustration of the counting predictor and line predictor. The former takes high-level features and predicted point maps as input to perform global plate number regression. The latter integrates low-level features and point predictions to produce line maps.}
	\label{fig:counting_line_predictor}
\end{figure}
\subsection{Multi-Dimensional Decoder}
Our multi-dimensional decoder consists of a point predictor, counting predictor, and line predictor. The point predictor adopts a general U-shaped architecture~\cite{Unet,FPN}, progressively fusing features output by each PFSSM to generate a coarse point map highly responsive to point locations. These coarse anode and cathode maps are subsequently used by both the counting and line predictors. 
As shown in Fig.~\ref{fig:counting_line_predictor}, it illustrates counting predictor and line predictor.

\noindent$\bullet$~\textit{Counting predictor.}
The overall counting predictor is constructed to solve a regression problem. The inputs of the counting predictor are the high-level features ($F_{current}^{e5}$)  and the predicted 
 point maps ($M^{anode}_{p}$, $M^{cathode}_{p}$) of the anode and cathode plates from the point predictor. To narrow the search range of the counting objects, we utilize two predicted point maps as spatial attention to guide $F_{current}^{e5}$. The number of anode and cathode plates are computed as:
\begin{equation}\label{equ:1}
\centering
\left\{\begin{matrix}
    N^{anode} = \texttt{ReLU}(\texttt{Conv}(\texttt{GAP}(\texttt{DS}(F_{current}^{e5}) \otimes M^{anode}_{p})))\\
 N^{cathode} = \texttt{ReLU}(\texttt{Conv}(\texttt{GAP}(\texttt{DS}(F_{current}^{e5}) \otimes M^{cathode}_{p}))),\\
    \end{matrix}\right.
\end{equation} 
where \texttt{DS(·)} is downsampling, $\otimes$ is element-wise multiplication, \texttt{GAP(·)} is global average pooling, and \texttt{Conv(·)} is a convolution layer with one-channel output. 
As an auxiliary task for the point segmentation, counting task can improve the query ability of high-level features on the number of plates at the global level, thereby enhancing the feature representation of the point branch.

\noindent$\bullet$~\textit{Line predictor.}
Generally speaking, detail information of contours often exists in the low-level features. Therefore, we build the line predictor with both low-level features ($F_{current}^{e1}$, $F_{current}^{e2}$) and the predicted 
 point maps ($M^{anode}_{p}$, $M^{cathode}_{p}$) as inputs. 
We first aggregate the $F_{current}^{e1}$, $F_{current}^{e2}$ to generate the $F_{current}^{e1,2}$:
\begin{equation}\label{equ:2}
\centering
\begin{matrix}
    F_{current}^{e1,2} = \texttt{Conv}(F_{current}^{e1} + \texttt{US}(F_{current}^{e2})), 
\end{matrix}
\end{equation} 
where \texttt{US(·)} is the upsampling operation. Next, $F_{current}^{e1,2}$ is separately computed with  $M^{anode}_{p}$ and $M^{cathode}_{p}$
in the form of residual~\cite{Resnet} and predict line segmentation maps:
\begin{equation}\label{equ:3}
\centering
\left\{\begin{matrix}
    L^{anode} = \texttt{S}(\texttt{Conv}(M^{anode}_{p} \otimes F_{current}^{e1,2}) + F_{current}^{e1,2})\\
  L^{cathode} = \texttt{S}(\texttt{Conv}(M^{cathode}_{p} \otimes F_{current}^{e1,2}) + F_{current}^{e1,2}),\\
    \end{matrix}\right.
\end{equation} 
where \texttt{S(·)} is the element-wise sigmoid. 
As another auxiliary task of point segmentation, line segmentation can provide continuous segmentation clues to compensate and modify some plates that are not accurately predicted by point branch. 

 \begin{figure}[!t]
	\centering
	\includegraphics[width=\linewidth]{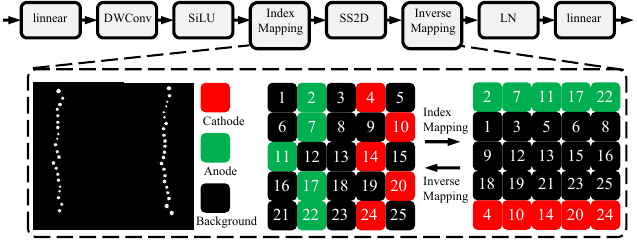}
	\caption{Illustration of the density-aware reordering state space module. It takes low-level features and a coarse point map as inputs.  Index mapping rearranges semantically similar pixels, and inverse mapping restores the original spatial layout after SS2D.
}
	\label{fig:DRSM}
\end{figure}

\subsection{Density-aware Reordering State Space Module}
{
As shown in Fig.~\ref{fig:DRSM}, the proposed Density-aware Reordering State Space Module (DRSSM) aims to enhance long-range dependency modeling for densely arranged electrode plates by explicitly organizing spatial tokens according to their semantic density. DRSSM takes the low-level feature and the coarse point prediction map generated by the point predictor as inputs.}

\noindent{\noindent$\bullet$~\textit{Feature projection.}
Given the input feature map $F \in \mathbb{R}^{H \times W \times C}$, we first project it through a linear layer, followed by depthwise separable convolution (\texttt{DWConv}) and a \texttt{SiLU} activation to enhance local spatial representation while maintaining computational efficiency.}

\noindent{\noindent$\bullet$~\textit{Semantic map generation.}
Let $\hat{P} \in \mathbb{R}^{H \times W}$ denote the coarse point prediction map produced by the point predictor, which encodes coarse semantic categories such as anode, cathode, and background. To suppress prediction noise and stabilize the subsequent token ordering, $\hat{P}$ is downsampled using max pooling with a kernel size equal to the patch size. A fixed threshold $\tau$ (set to $0.5$ in our implementation) is then applied to convert the pooled map into discrete semantic labels for each spatial location.}

\noindent{\noindent$\bullet$~\textit{Density-aware index generation and token reordering.}
Based on the discrete semantic labels, we construct a per-image semantic index vector by flattening the 2D label map. The feature map $F$ is reshaped into a sequence representation and reordered by sorting this semantic index vector, producing a reordered sequence $F_{\text{reordered}} \in \mathbb{R}^{N \times C}$, where tokens belonging to the same semantic category are grouped together. This density-aware rearrangement brings semantically consistent tokens into closer sequential proximity, which facilitates more coherent dependency modeling.}

\noindent{\noindent$\bullet$~\textit{State space processing.}
The reordered sequence is then fed into the SS2D module, where selective scanning is performed along the reordered dimension. By operating on semantically grouped tokens, the state space model emphasizes intra-class continuity and reduces interference between foreground plates and background regions. This design enables more reliable modeling of plate structures under dense arrangements and ambiguous boundaries.}

\noindent{\noindent$\bullet$~\textit{Inverse mapping and spatial reconstruction.}
After state space processing, the output sequence is restored to its original spatial layout using the inverse index mapping saved during the reordering step. This inverse projection ensures that each token is placed back to its correct spatial position, yielding a refined feature map
$F_{\text{refined}} \in \mathbb{R}^{H \times W \times C}$. 
It is subsequently normalized using \texttt{LayerNorm} and projected through a linear layer to produce the final refined point-wise prediction, which improves plate boundary discrimination and instance separability.}

\begin{figure}[t]
\centering
\includegraphics[width=1.0\linewidth]{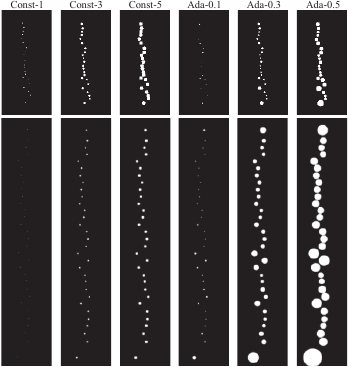}
\caption{Visualization of the ground-truth point masks under different strategies for point label generation.}
\label{fig:point_mask_strategy}
\end{figure}

\subsection{Label Generation and Supervision}\label{sec:generation_mask}
For point segmentation, a direct strategy of generating the ground truth of point mask is to use circular regions of the same radius for each plate. As shown in the $1^{st}$ - $3^{rd}$ columns of Fig.~\ref{fig:point_mask_strategy}, the point masks are yielded under the radius of $1$, $3$, and $5$ pixels, respectively. It can be seen that the fixed radius often produces   very dense or sparse point masks for different types of batteries, which increases the difficulty of point segmentation. To this end, we design a novel distance-adaptive label generation strategy. It computes the diameter of each point according to the distance of adjacent plates. As shown in the $4^{th}$ - $6^{th}$ columns of Fig.~\ref{fig:point_mask_strategy}, the point masks are generated under the diameter of $0.1x$, $0.3x$, and $0.5x$ distance, respectively. The labels of counting and line segmentation can be separately obtained by the connected component analysis and connecting each endpoint in the point mask in sequence.
For the point and line segmentation predictors, we use the weighted IoU loss and binary cross entropy (BCE) loss, which have been widely adopted in segmentation tasks. We use the same definitions as in~\cite{SPNet,F3Net,PraNet,MSNet_Polyp}. For the counting predictor, we adopt the common L1 loss. 
The total training loss is written as follows:
\begin{equation}\label{equ:loss}
\mathcal{L}_{\text{total}} = \lambda_1 \cdot \mathcal{L}_{\text{point}}^{\text{refine}} 
+ \lambda_2 \cdot \mathcal{L}_{\text{point}}^{\text{coarse}} 
+ \lambda_3 \cdot \mathcal{L}_{\text{count}} 
+ \lambda_4 \cdot \mathcal{L}_{\text{line}}, 
\end{equation}
where \( \lambda_1, \lambda_2, \lambda_3, \lambda_4 \) are  balancing weights, empirically set to  \( \lambda_1 = \lambda_2 = 1 \), \( \lambda_3 = 0.05 \), and \( \lambda_4 = 0.5 \). Point segmentation receives the highest weights as it is our core task, while count loss is down-weighted to avoid dominating early optimization due to its large magnitude.

\section{Experiments}
\subsection{Implementation Details}
Our MDCNeXt model is implemented in PyTorch and trained on 4 NVIDIA Tesla V100 GPUs for 150 epochs with a batch size of 4. All X-ray images are resized to $512 \times 512$. We adopt the Adam optimizer~\cite{Adam} with $\beta_1 = 0.5$, $\beta_2 = 0.999$. The initial learning rate is set to $1 \times 10^{-4}$. To stabilize training, we apply a weight decay of $1 \times 10^{-3}$ and gradient clipping with a threshold of 0.5. A step-wise learning rate decay schedule reduces the learning rate by 0.9 every 120 epochs. To improve generalization, we apply data augmentation techniques including random horizontal flipping, multi-scale resizing (0.75$\times$, 1.0$\times$, 1.25$\times$), and random brightness adjustment.
To prevent the model from overfitting to a specific prompt during inference,  we adopt a prompt randomization strategy throughout training. Instead of using a fixed prompt sample, each training epoch randomly selects a pure plate from the training set to serve as the prompt input. This approach encourages the model to learn robust prompt-invariant representations.  {We explicitly detail the experimental settings for each category of baselines to ensure transparency and fairness in the comparisons. The corresponding protocols are summarized as follows.}

{1) Corner Detection Methods.
Traditional corner detection methods are classical image processing approaches and do not require learning or training. To reduce the influence of severe noise in industrial X-ray images and ensure reasonable performance, we provide these methods with Canny edge detection results as input, which is a standard preprocessing practice in industrial vision and does not introduce learnable bias.}

{2) Crowd Counting Methods.
All crowd counting baselines are re-trained on PBD5K by strictly following the official open-source implementations and training configurations reported in their original papers, including optimizer settings, learning rates, and training schedules.}

{3) General and Tiny Object Detection Methods.
Similarly, all general-purpose and tiny object detection models are re-trained on PBD5K using their official codebases and default training protocols, without additional task-specific tuning beyond dataset adaptation.}

{4) Segmentation Methods.
The segmentation baselines are divided into three categories:}
\begin{itemize}
    \item 
    {Prompt-based generalist models (SAM 2~\cite{SAM2} and SegGPT~\cite{SegGPT}): They are pre-trained on extremely large-scale generic segmentation datasets and are not re-trained or fine-tuned on PBD5K. To evaluate their capability to cover the PBD task under their intended usage paradigm, we adopt the same pure battery images and corresponding masks used in MDCNeXt as prompts, and directly generate predictions on the test set. This protocol follows the standard inference setting recommended by these models and ensures a fair prompt-level comparison.}
    \item 
    {General segmentation frameworks (DeepLabV3+~\cite{DeepLabV3+} and  SegFormer~\cite{SegFormer}): These models are re-trained on PBD5K using their official open-source code and training configurations, with identical training data splits and supervision.}
    \item 
     {Context-dependent concept segmentation models (Spider~\cite{CDCU-Spider} and ZoomNeXt~\cite{ZoomNeXt}): These models are also re-trained on PBD5K under the same training protocols as their original implementations, ensuring consistent optimization and data exposure.}
\end{itemize}

\begin{table*}[!t]
\centering
	\caption{Quantitative comparison of different methods. $\uparrow$ and $ \downarrow$ indicate that the larger scores and the smaller ones  are better, respectively. The best scores are highlighted in {\color{reda} \textbf{red}}. ``—'' represents that the results are not available because these methods can not provide coordinate information or their prediction accuracy of the number of plates is zero. ``Average'' refers to the average scores across the Regular, Difficult, and Tough test splits.}
   \resizebox{\linewidth}{!}{
    \setlength\tabcolsep{2pt}
    \renewcommand\arraystretch{1.2}
  \begin{tabular}{c|r||ccc|ccc|ccccc|cc}
  	\toprule[2pt]
  	&  & \multicolumn{3}{c|}{\textbf{{Corner Detection}}}& \multicolumn{3}{c|}{\textbf{{Crowd Counting }}}& \multicolumn{5}{c|}{\textbf{{General/Tiny Object Detection
  	}}} & \multicolumn{2}{c}{\textbf{{Segmentation
  	}}} \\
  	Dataset&Metrics  & Harris & Shi-tomasi
  	&Sub-pixel& BL & CUT & IOCFormer & RetinaNet & DETR
  	&YOLOv10&CFINet&HS-FPN&MDCNet &MDCNeXt 
  	\\
  	&  & ~\cite{Harris} & ~\cite{Shitomasi}
  	&~\cite{Sub-Pixel}& ~\cite{BL} & ~\cite{CUT_cc} & ~\cite{IOCFormer} & ~\cite{RetinaNet} &~\cite{DETR}
  	&~\cite{Yolov10}&~\cite{C3Det}&~\cite{HS-FPN} &CVPR 2024~\cite{MDCNet}&(Ours)
  	\\
  	\hline
  	\multirow{8}{*}{Regular} 
  	&AN-MAE$\downarrow$
  	&195.2021
  	&121.4027
  	&165.5050
  	&3.8430
  	&2.9021
  	&2.3879
  	&	2.1290 
  	&	3.2294 
  	&	2.7320 
  	&	1.9377 
  	&	1.3402  
  	&	 {0.3028}
  	&	\color{reda}{\textbf{0.2175}}$_{\text{\textcolor{mygreen}{$\uparrow$28\%}}}$ 
  	\\
  	&CN-MAE$\downarrow$
  	&4892.2022
  	&239.3201
  	&219.2020
  	&3.1252
  	&2.3026
  	&2.0352
  	&	1.9346 
  	&	5.3212 
  	&	4.3824 
  	&	1.3798 
  	&	1.1202  
  	&{0.2187}
  	&\color{reda}\textbf{0.1320}$_{\text{\textcolor{mygreen}{$\uparrow$40\%}}}$ 
  	\\
  	&AN-ACC$\uparrow$
  	&0.0000
  	&0.0000
  	&0.0000
  	&0.4892
  	&0.5324
  	&0.4512
  	&0.6483
  	&0.5186
  	&0.6232
  	&0.6825
  	&0.7024
  	&{0.8932}
  	&\color{reda}\textbf{0.9379}$_{\text{\textcolor{mygreen}{$\uparrow$5\%}}}$ 
  	\\
  	&CN-ACC$\uparrow$
  	&0.0000
  	&0.0000
  	&0.0000
  	&0.5112
  	&0.5512
  	&0.4732
  	&0.6626
  	&0.4984
  	&0.5465
  	&0.6942
  	&0.7360
  	&{0.8964}
  	&\color{reda}\textbf{0.9456}$_{\text{\textcolor{mygreen}{$\uparrow$5\%}}}$ 
  	\\
  	&PN-ACC$\uparrow$
  	&0.0000
  	&0.0000
  	&0.0000
  	&0.4673
  	&0.4956
  	&0.4072
  	&0.6135
  	&0.4215
  	&0.4576
  	&0.6356
  	&0.6735
  	&{0.8531}
  	&\color{reda}\textbf{0.9165}$_{\text{\textcolor{mygreen}{$\uparrow$7\%}}}$ 
  	\\
  	&AL-MAE$\downarrow$
  	&--
  	&--
  	&--
  	&--
  	&--
  	&--
  	&4.5640
  	&8.5600
  	&3.4985
  	&2.4745
  	&2.3475
  	&{1.3245}
  	&\color{reda}\textbf{1.1537}$_{\text{\textcolor{mygreen}{$\uparrow$13\%}}}$ 
  	\\
  	&CL-MAE$\downarrow$
  	&--
  	&--
  	&--
  	&--
  	&--
  	&--
  	&4.7754
  	&7.9456
  	&3.0782
  	&2.3750
  	&2.1576
  	&{1.3948}
  	&\color{reda}\textbf{1.2234}$_{\text{\textcolor{mygreen}{$\uparrow$12\%}}}$ 
  	
  	\\
  	&OH-MAE$\downarrow$
  	&--
  	&--
  	&--
  	&--
  	&--
  	&--
  	&3.9432
  	&8.2468
  	&2.9670
  	&2.4150
  	&2.3756
  	&{1.3750}
  	&\color{reda}\textbf{1.0102}$_{\text{\textcolor{mygreen}{$\uparrow$27\%}}}$ 
  	\\
  	\hline
  	\multirow{8}{*}{Difficult} 
  	&AN-MAE$\downarrow$
  	&136.7474
  	&84.1242
  	&115.5250
  	&1.7576
  	&2.6755
  	&1.4256
  	&1.7576
  	&2.8465
  	&2.0789
  	&2.8645
  	&1.3974
  	&{0.4732}
  	&\color{reda}\textbf{0.3515}$_{\text{\textcolor{mygreen}{$\uparrow$26\%}}}$ 
  	\\
  	&CN-MAE$\downarrow$
  	&4218.2002
  	&178.1542
  	&245.0121
  	&1.8546
  	&3.0250
  	&1.6251
  	&1.7450
  	&3.0856
  	&2.0227
  	&2.9875
  	&1.3746
  	&{0.4450}
  	&\color{reda}\textbf{0.3428}$_{\text{\textcolor{mygreen}{$\uparrow$23\%}}}$ 
  	
  	\\
  	&AN-ACC$\uparrow$
  	&0.0000
  	&0.0000
  	&0.0000
  	&0.3986
  	&0.2578
  	&0.3659
  	&0.4745
  	&0.3145
  	&0.4576
  	&0.5175
  	&0.5846
  	&{0.8078}
  	&\color{reda}\textbf{0.8973}$_{\text{\textcolor{mygreen}{$\uparrow$11\%}}}$ 
  	\\
  	&CN-ACC$\uparrow$
  	&0.0000
  	&0.0000
  	&0.0000
  	&0.2546
  	&0.1974
  	&0.4745
  	&0.3610
  	&0.3091
  	&0.3420
  	&0.4010
  	&0.4715
  	&{0.7746} 
  	&\color{reda}\textbf{0.8552}$_{\text{\textcolor{mygreen}{$\uparrow$10\%}}}$ 
  	
  	\\
  	&PN-ACC$\uparrow$
  	&0.0000
  	&0.0000
  	&0.0000
  	&0.2847
  	&0.1127
  	&0.2461
  	&0.2587
  	&0.2545
  	&0.2475
  	&0.2978
  	&0.3726
  	&{0.6988}
  	&\color{reda}\textbf{0.7983}$_{\text{\textcolor{mygreen}{$\uparrow$14\%}}}$ 
  	
  	\\
  	&AL-MAE$\downarrow$
  	&--
  	&--
  	&--
  	&--
  	&--
  	&--
  	&10.2643
  	&31.4544
  	&9.1798
  	&8.9599
  	&8.4756
  	&{4.1547}
  	&\color{reda}\textbf{1.3702}$_{\text{\textcolor{mygreen}{$\uparrow$67\%}}}$ 
  	
  	\\
  	&CL-MAE$\downarrow$
  	&--
  	&--
  	&--
  	&--
  	&--
  	&--
  	&10.9875
  	&34.7055
  	&9.3256
  	&8.8750
  	&8.2745
  	&{4.0025} 
  	&\color{reda}\textbf{1.2953}$_{\text{\textcolor{mygreen}{$\uparrow$68\%}}}$ 
  	
  	\\
  	&OH-MAE$\downarrow$
  	&--
  	&--
  	&--
  	&--
  	&--
  	&--
  	&6.0875
  	&12.2540
  	&7.2560
  	&7.1515
  	&7.4250
  	&{3.9846}
  	&\color{reda}\textbf{1.1603}$_{\text{\textcolor{mygreen}{$\uparrow$71\%}}}$ 
  	\\
  	\hline
  	\multirow{8}{*}{Tough} 
  	&AN-MAE$\downarrow$
  	&275.4540
  	&165.1059
  	&190.0571
  	&3.5425
  	&2.9546
  	&2.8472
  	&7.1502
  	&12.3564
  	&8.7600
  	&7.4658
  	&6.9875
  	&{4.2547}
  	&\color{reda}\textbf{0.7873}$_{\text{\textcolor{mygreen}{$\uparrow$81\%}}}$ 
  	
  	\\
  	&CN-MAE$\downarrow$
  	&4750.1010
  	&225.0325
  	&295.3512
  	&2.1978
  	&2.1785
  	&3.0201
  	&6.3572
  	&9.0971
  	&11.1015
  	&7.1150
  	&6.0220
  	&{3.1042}
  	&\color{reda}\textbf{0.3781}$_{\text{\textcolor{mygreen}{$\uparrow$88\%}}}$ 
  	
  	\\
  	&AN-ACC$\uparrow$
  	&0.0000
  	&0.0000
  	&0.0000
  	&0.2512
  	&0.3741
  	&0.3075
  	&0.1912
  	&0.0975
  	&0.2541
  	&0.2987
  	&0.3105
  	&{0.5402}
  	&\color{reda}\textbf{0.7873}$_{\text{\textcolor{mygreen}{$\uparrow$46\%}}}$ 
  	\\
  	&CN-ACC$\uparrow$
  	&0.0000
  	&0.0000
  	&0.0000
  	&0.2843
  	&0.2915
  	&0.4253
  	&0.3102
  	&0.1423
  	&0.1967
  	&0.2547
  	&0.3746
  	&{0.5015}
  	&\color{reda}\textbf{0.7341}$_{\text{\textcolor{mygreen}{$\uparrow$46\%}}}$ 
  	
  	\\
  	&PN-ACC$\uparrow$
  	&0.0000
  	&0.0000
  	&0.0000
  	&0.2034
  	&0.2545
  	&0.2275
  	&0.1340
  	&0.0672
  	&0.0975
  	&0.1440
  	&0.2116
  	&{0.4125}
  	&\color{reda}\textbf{0.6263}$_{\text{\textcolor{mygreen}{$\uparrow$52\%}}}$ 
  	\\
  	&AL-MAE$\downarrow$
  	&--
  	&--
  	&--
  	&--
  	&--
  	&--
  	&17.8345
  	&48.2451
  	&20.0240
  	&14.3545
  	&17.1545
  	&{8.3425}
  	&\color{reda}\textbf{1.2145}$_{\text{\textcolor{mygreen}{$\uparrow$85\%}}}$ 
  	\\
  	&CL-MAE$\downarrow$
  	&--
  	&--
  	&--
  	&--
  	&--
  	&--
  	&20.9871
  	&54.0202
  	&22.1544
  	&16.2540
  	&14.8741
  	&{7.5471}
  	&\color{reda}\textbf{0.9826}$_{\text{\textcolor{mygreen}{$\uparrow$87\%}}}$ 
  	\\
  	&OH-MAE$\downarrow$
  	&--
  	&--
  	&--
  	&--
  	&--
  	&--
  	&7.0470
  	&18.9345
  	&8.1574
  	&9.0202
  	&8.1345
  	&{5.3451}
  	&\color{reda}\textbf{0.9979}$_{\text{\textcolor{mygreen}{$\uparrow$81\%}}}$ 
  	\\
  	\hline \hline
  	\multirow{8}{*}{Average} 
  	&AN-MAE$\downarrow$
  	& 198.7517  &121.1357 & 153.6240  &   2.8988 &   2.8283 &   2.1546 &   3.6786
  	&  6.1642  &  4.5086  &  4.1834 &   3.2749 &   1.7094     &\color{reda}\textbf{0.4645}$_{\text{\textcolor{mygreen}{$\uparrow$73\%}}}$ 
  	
  	\\
  	&CN-MAE$\downarrow$
  	&  4571.8041  & 209.7727 &  255.4058  &   2.2980  &    2.5524  &   2.2029 &    3.3551
  	&   5.6962  &   5.7035  &   3.9707  &   2.8822  &   1.2869  &   \color{reda}\textbf{0.3005}$_{\text{\textcolor{mygreen}{$\uparrow$77\%}}}$ 
  	
  	\\
  	&AN-ACC$\uparrow$
  	& 0.0000      &  0.0000    &     0.0000   &      0.3720  &    0.3679    & 0.3681     &0.4234
  	&0.2936    & 0.4314    & 0.4859   &  0.5222     &0.7392    & \color{reda}\textbf{0.8705}$_{\text{\textcolor{mygreen}{$\uparrow$18\%}}}$ 
  	\\
  	&CN-ACC$\uparrow$
  	&  0.0000       &   0.0000    &      0.0000   &       0.3307  &    0.3204  &    0.4575    &  0.4215
  	&    0.3014  &    0.3455  &    0.4270  &     0.5068  &    0.7135   &   \color{reda}\textbf{0.8375}$_{\text{\textcolor{mygreen}{$\uparrow$17\%}}}$ 
  	
  	\\
  	&PN-ACC$\uparrow$
  	&0.0000      &  0.0000    &    0.0000  &      0.3042 &   0.2593  &  0.2813    &0.3079
  	& 0.2341  &  0.2508 &   0.3327 &   0.3956   & 0.6416   & \color{reda}\textbf{0.7705}$_{\text{\textcolor{mygreen}{$\uparrow$20\%}}}$ 
  	\\
  	&AL-MAE$\downarrow$
  	&--       &--        &--      &--      &--      &--      &11.3590
  	&31.2427    &11.3876     &9.1160      &9.8354    & 4.8435    & \color{reda}\textbf{1.2617}$_{\text{\textcolor{mygreen}{$\uparrow$74\%}}}$ 
  	\\
  	&CL-MAE$\downarrow$
  	&--       &--        &--      &--      &--      &--       &12.7727
  	& 34.3529   &12.0594   & 9.6990    & 8.9334  &  4.5309  &  \color{reda}\textbf{1.1709}$_{\text{\textcolor{mygreen}{$\uparrow$74\%}}}$ 
  	\\
  	&OH-MAE$\downarrow$
  	&--       &--        &--      &--      &--      &--          & 5.8601
  	& 13.4835   & 6.4567   & 6.5644&    6.3649   & 3.7732   & \color{reda}\textbf{1.0667}$_{\text{\textcolor{mygreen}{$\uparrow$72\%}}}$ 
  	\\
  	\bottomrule[2pt]
  \end{tabular}
	}
	\label{tab:comparison}
\end{table*}

\begin{figure*}[!t]
\centering
\includegraphics[width=0.99\linewidth]{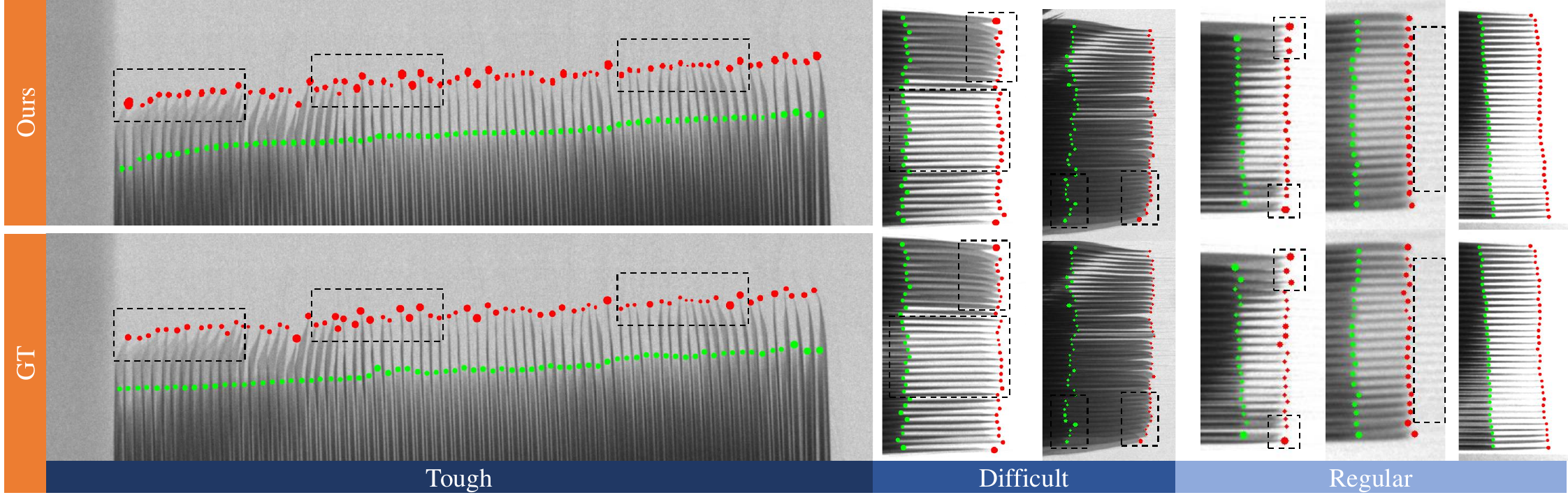}
\caption{Qualitative results on regular, difficult, and tough examples with varying attributes, resolutions, overhang levels, and  numbers of plates. Black dashed boxes (\texttt{[--]})  highlight the challenging regions. Best viewed zoomed in.}
\label{fig:Visual_results} 
\end{figure*}

\subsection{Comparison with Corner Detection, Crowd Counting, General and Tiny Object Detection Solutions}
\textbf{Quantitative Comparison.}  
As shown in Tab.~\ref{tab:comparison}, we conduct a comprehensive comparison across various representative methods that serve as potential solutions to the PBD task. The evaluation spans regular, difficult, and tough splits, and we further report the average scores across them. The key observations are as follows: 
\textit{{\uppercase\expandafter{\romannumeral1})}}  Our segmentation-based MDCNeXt and MDCNet achieve state-of-the-art performance across different splits under all metrics. 
It confirms that modeling the PBD task as a visual point segmentation problem is a correct solution.  
Moreover, MDCNeXt further outperforms MDCNet
with average gains exceeding 18\%, 17\%, and 20\% on AN-ACC, CN-ACC, and PN-ACC, respectively. 
\textit{{\uppercase\expandafter{\romannumeral2})}}  The complementarity among different evaluation metrics highlights the necessity of using a multi-metric evaluation scheme for PBD. 
For example, tiny object detection methods have AN-MAE and CN-MAE scores close to ours (within 2 units) on the regular split. However, their AN-ACC and CN-ACC drop by more than 20\% compared to MDCNeXt. Besides, the OH-MAE metric may fail to capture systematic shifts. 
When both AN and CN position predictions are slightly displaced in the same direction, OH-MAE might still appear low. In the tough split, other methods show AL-MAE and CL-MAE scores that are more than twice their OH-MAE, exposing serious localization issues. Therefore, combining multiple metrics gives a more reliable and precise evaluation. This is consistent with the strict spatial accuracy requirements of the PBD task, where even small errors are magnified and ``almost correct'' results are not acceptable.
\textit{{\uppercase\expandafter{\romannumeral3})}}  MDCNeXt shows strong robustness and stability across all splits. Its AL-/CL-MAE values fluctuate within only 0.3. In contrast, other object detection methods suffer from significant accuracy degradation as the number of plates sheets increases and the accumulation of interference types. In the tough split, their AL-MAE and CL-MAE scores exceed 15, reflecting a strong decline in positional precision compared to the regular split. 

{
\textbf{Low Performance on Tough Split.} 
We can observe that MDCNeXt, like other methods, experiences a noticeable performance drop in number accuracy when facing the tough split. 
It is important to note that our number accuracy (AN/CN/PN-ACC) metric is extremely strict. This metric is defined \textbf{at the sample level}, meaning that for a given image, if the predicted number of plates does not perfectly match the ground truth, even by a single plate, the ACC for that image is recorded as zero. As a result, images containing very dense and numerous plates naturally have a higher risk of minor counting deviations, which can disproportionately reduce the reported ACC.
If the evaluation is instead conducted \textbf{at the plate level}, our MDCNeXt can achieve an PN-ACC of up to 96\% on the Tough split. 
The noticeable gap between plate-level and sample-level ACC reflects the intrinsic difficulty of the tough split, where extreme plate density, severe plate adhesion, and weak inter plate contrast substantially increase the likelihood of non perfect predictions. This gap should be interpreted as a consequence of the stringent evaluation protocol rather than a failure of the model’s segmentation capability.
In real industrial inspection pipelines, plate number prediction is not a single standalone decision point. Low-confidence or contradictory predictions (e.g., large deviation from the expected plate count defined by the cell design) naturally trigger re-inspection mechanisms, including manual sampling and upstream process checks. In practice, such failures are more likely to serve as alarms indicating potential process drift or manufacturing defects, rather than silently propagating incorrect decisions. Moreover, from an industrial safety perspective, metrics related to overhang are often more critical than absolute plate count accuracy, as overhang directly correlates with short-circuit and explosion risks.
}

\begin{figure*}[t]
\centering
\includegraphics[width=1.0\linewidth]{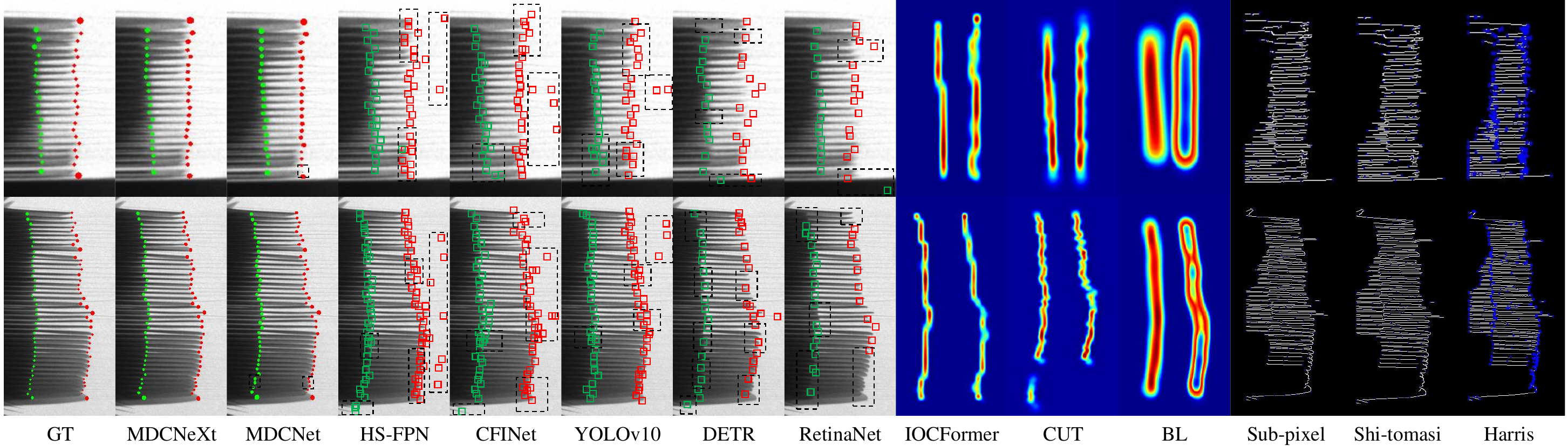}
\caption{Visual comparison with other general/tiny object detection-based~\cite{HS-FPN,CFINet,Yolov10,DETR,RetinaNet}, counting-based~\cite{BL,CUT_cc,IOCFormer}, corner detection-based~\cite{Sub-Pixel,Shitomasi,Harris} solutions. We directly visualize the predicted results (Ours: Segmentation map, General/Tiny object detection methods: Bounding box, Counting methods: Density map, Corner detection methods: Corner map) without any post-processing operations. Black dashed boxes (\texttt{[--]}) highlight the regions with obvious prediction failures. Best viewed zoomed in.}
\label{fig:other_compar_mdcnext}
\end{figure*}

\begin{table*}[!t]
\centering 
    \caption{Quantitative comparison of image segmentation methods. All ten metrics are widely used across various segmentation branches~\cite{SegFormer,ZoomNet,PraNet,CDCU-Spider}, with details introduced in the survey~\cite{GateNetv2}.  
$\uparrow$ and $\downarrow$ indicate that higher and lower values are better, respectively. Best scores are highlighted in  {\color{reda} \textbf{red}}. 
MDCNeXt's relative improvements over MDCNet are highlighted in  {\color{mygreen} \textbf{green}}. 
``Average'' denotes the mean scores across Regular, Difficult, and Tough test splits. For prompt-based generalist models (\textit{i.e.}, SAM 2 and SegGPT), we use the same pure battery images and their corresponding masks as those in MDCNeXt as prompts to generate predictions for the test set.
}
   \resizebox{0.9\linewidth}{!}{
    \setlength\tabcolsep{2pt}
    \renewcommand\arraystretch{1.2}
   \begin{tabular}{c|r|r|r||c|c|c|c|c|c|c|c|c|c}
   	\toprule[2pt]
   	Dataset&Method  & Pub. & Backbone& $PA\uparrow$ 	& $F_{\beta}^{max}\uparrow$ & $F_{\beta}^{mean}\uparrow$ & $F_{\beta}^{\omega}\uparrow$ & $S_m\uparrow$ & $E_m\uparrow$  &$mIoU\uparrow$&$mDice\uparrow$  &$BER\downarrow$ &$\mathcal{M}\downarrow$ 
   	\\
   	\hline
   	
   	\multirow{6}{*}{Regular}  
   	&SAM 2~\cite{SAM2}
   	&ICLR 2025&Hiera-B+~\cite{Hiera}
   	& 0.9802  &0.2473 &0.2213 &0.1979 &0.5932 &0.7026 &   0.5121&   0.2722&  0.3541&0.0177 
   	
   	\\
   	&SegGPT~\cite{SegGPT}&ICCV 2023 & ViT-L~\cite{ViT} 
   	& 0.9818  &0.2163 &0.1958 &0.1932 &0.5541 &0.7319 &   0.5551&   0.1958&  0.3880&0.0182 
   	
   	\\
   	&DeepLabV3+~\cite{DeepLabV3+}
   	&ECCV 2018&ResNet-101~\cite{Resnet}
   	& 0.9867  & 0.5078&  0.4278 & 0.4106 & 0.6438 & 0.9124 &  0.5920&  0.4379& 0.2711&0.0147 
   	
   	\\
   	&SegFormer~\cite{SegFormer}
   	&NeurIPS 2021&MiT-B5~\cite{SegFormer}
   	& 0.9848  & 0.5435&  0.4326 & 0.4020 & 0.6726 & 0.9074 &  0.6020&  0.4541& 0.2820&0.0134

   	\\
   	&Spider~\cite{CDCU-Spider}
   	&ICML 2024&ConvNeXt-L~\cite{ConvNeXt}
   	& 0.9888  & 0.6425&  0.5970 & 0.6212 & 0.7836 & 0.9425 &  0.6820&  0.6278& 0.1422&0.0097 
   	\\
   	&ZoomNeXt~\cite{ZoomNeXt}
   	&TPAMI 2024&ResNet-50~\cite{Resnet}
   	& 0.9888  & 0.6320&  0.6073 & 0.6320 & 0.7825 & 0.9617 &  0.6923&  0.6342& 0.1302&0.0092 
   	\\
   	\cline{2-14}
   	&MDCNet~\cite{MDCNet}
   	&CVPR 2024&ResNet-50~\cite{Resnet} 
   	& 0.9909  & 0.6730&  0.6635 & 0.7083 & 0.7912 & 0.9687 &  0.7432&  0.7023& 0.1107&0.0081 
   	\\
   	&MDCNeXt
   	&--&ResNet-50~\cite{Resnet} 
   	& \color{reda}\textbf{0.9924}  & \color{reda}\textbf{0.6979}&  \color{reda}\textbf{0.6971} &\color{reda}\textbf{0.7337} & \color{reda}\textbf{0.8113} & \color{reda}\textbf{0.9786} &  \color{reda}\textbf{0.7857}&  \color{reda}\textbf{0.7248}& \color{reda}\textbf{0.1101}&\color{reda}\textbf{0.0076}  \\
   	&    &  & 
   	& \color{mygreen}{$\uparrow$0.2\%} 
   	& \color{mygreen}{$\uparrow$3.7\%} 
   	& \color{mygreen}{$\uparrow$5.1\%} 
   	&\color{mygreen}{$\uparrow$3.6\%} 
   	&\color{mygreen}{$\uparrow$2.5\%}
   	&\color{mygreen}{$\uparrow$1.0\%}
   	&\color{mygreen}{$\uparrow$5.7\%}
   	&\color{mygreen}{$\uparrow$3.2\%}
   	&\color{mygreen}{$\uparrow$0.5\%}
   	&\color{mygreen}{$\uparrow$6.2\%}
   	\\
   	\hline
   	\multirow{6}{*}{Difficult} 
   	&SAM 2~\cite{SAM2}
   	&ICLR 2025&Hiera-B+~\cite{Hiera}
   	& 0.9836  & 0.1989&  0.1782 & 0.1678 & 0.5029 & 0.7302 &  0.5780&  0.1011& 0.4474&0.0187 
   	
   	\\
   	&SegGPT~\cite{SegGPT}&ICCV 2023 & ViT-L~\cite{ViT} 
   	& 0.9817  &0.1499 &0.1359& 0.1337 &0.5281& 0.6867 &   0.5335&  0.1347& 0.4233 &0.0182 
   	
   	\\
   	&DeepLabV3+~\cite{DeepLabV3+}
   	&ECCV 2018&ResNet-101~\cite{Resnet}
   	& 0.9865  & 0.3691&  0.3742 & 0.3542 & 0.4109 & 0.9102 &  0.5532&  0.3801& 0.3080&0.0142 
   	
   	\\
   	&SegFormer~\cite{SegFormer}
   	&NeurIPS 2021&MiT-B5~\cite{SegFormer}
   	& 0.9888  & 0.3390&  0.3512 & 0.3210 & 0.3902 & 0.8921 &  0.5721&  0.3670& 0.3303&0.0153

   	\\
   	&Spider~\cite{CDCU-Spider}
   	&ICML 2024&ConvNeXt-L~\cite{ConvNeXt}
   	& 0.9900  & 0.4412& 0.4740 &0.4412 & 0.6021 & 0.9314 & 0.6023&  0.4921&  0.2344&0.0129 
   	\\
   	&ZoomNeXt~\cite{ZoomNeXt}
   	&TPAMI 2024&ResNet-50~\cite{Resnet}
   	& 0.9888  & 0.5012& 0.5079 & 0.4936 & 0.6785 & 0.9508 & 0.6289&  0.5182&  0.2161&0.0117 
   	\\
   	\cline{2-14}
   	&MDCNet~\cite{MDCNet}
   	&CVPR 2024&ResNet-50~\cite{Resnet} 
   	& 0.9908  & 0.5652& 0.5420 & 0.5372 & 0.7084 & 0.9538 & 0.6709&  0.5690&  0.1761&0.0094 
   	\\
   	&MDCNeXt
   	&--&ResNet-50~\cite{Resnet} 
   	& \color{reda}\textbf{0.9913}  & \color{reda}\textbf{0.6252}&  \color{reda}\textbf{0.6245} 
   	&\color{reda}\textbf{0.6611} & \color{reda}\textbf{0.7726} & \color{reda}\textbf{0.9758} &  \color{reda}\textbf{0.7459}&  \color{reda}\textbf{0.6522}& 
   	\color{reda}\textbf{0.1461}&\color{reda}\textbf{0.0086} 
   	\\
   	&    &  & 
   	& \color{mygreen}{$\uparrow$0.1\%} 
   	& \color{mygreen}{$\uparrow$10.6\%} 
   	& \color{mygreen}{$\uparrow$15.2\%} 
   	&\color{mygreen}{$\uparrow$23.1\%} 
   	&\color{mygreen}{$\uparrow$9.1\%}
   	&\color{mygreen}{$\uparrow$2.3\%}
   	&\color{mygreen}{$\uparrow$11.2\%}
   	&\color{mygreen}{$\uparrow$14.6\%}
   	&\color{mygreen}{$\uparrow$17.0\%}
   	&\color{mygreen}{$\uparrow$8.5\%}
   	\\
   	\hline
   	\multirow{6}{*}{Tough} 
   	&SAM 2~\cite{SAM2}
   	&ICLR 2025&Hiera-B+~\cite{Hiera}
   	& 0.9840  & 0.1312&  0.1102 & 0.0981 & 0.4980 & 0.7022 &  0.5320&  0.0921& 0.4467&0.0186 
   	
   	\\
   	&SegGPT~\cite{SegGPT}&ICCV 2023 & ViT-L~\cite{ViT} 
   	& 0.9819  & 0.0914 & 0.0824 & 0.0776&  0.5092&  0.6022  &  0.5158&  0.0790& 0.4579&0.0180
   	
   	\\
   	&DeepLabV3+~\cite{DeepLabV3+}
   	&ECCV 2018&ResNet-101~\cite{Resnet}
   	& 0.9878  & 0.2712& 0.1997 & 0.1939 & 0.2892 & 0.7920 & 0.3702&  0.3083&  0.3999&0.0178 
   	
   	\\
   	&SegFormer~\cite{SegFormer}
   	&NeurIPS 2021&MiT-B5~\cite{SegFormer}
   	& 0.9890  & 0.2521& 0.2001 & 0.2032 & 0.2782 & 0.7808 & 0.3840&  0.3012&  0.3911&0.0173

   	\\
   	&Spider~\cite{CDCU-Spider}
   	&ICML 2024&ConvNeXt-L~\cite{ConvNeXt}
   	& 0.9896  & 0.3831& 0.3221 & 0.3353 & 0.4483 & 0.9312 & 0.5201&  0.4032&  0.2411&0.0146  
   	\\
   	&ZoomNeXt~\cite{ZoomNeXt}
   	&TPAMI 2024&ResNet-50~\cite{Resnet}
   	& 0.9888  & 0.4378& 0.3792 & 0.3511 & 0.5189 & 0.9433 & 0.5811&  0.4732&  0.2235&0.0129 
   	\\
   	\cline{2-14}
   	&MDCNet~\cite{MDCNet}
   	&CVPR 2024&ResNet-50~\cite{Resnet} 
   	& 0.9890  & 0.4735& 0.4012 & 0.3832 & 0.5733 & 0.9511 & 0.6021&  0.5034&  0.2190&0.0121 
   	\\
   	&MDCNeXt
   	&--&ResNet-50~\cite{Resnet} 
   	& \color{reda}\textbf{0.9903}  & \color{reda}\textbf{0.5641}&  \color{reda}\textbf{0.5633} &\color{reda}\textbf{0.5793} & \color{reda}\textbf{0.7353} & \color{reda}\textbf{0.9745} &  \color{reda}\textbf{0.7153}&  \color{reda}\textbf{0.5872}& \color{reda}\textbf{0.1831}&
   	\color{reda}\textbf{0.0097} 
   	\\
   	&    &  & 
   	& \color{mygreen}{$\uparrow$0.1\%} 
   	& \color{mygreen}{$\uparrow$19.1\%} 
   	& \color{mygreen}{$\uparrow$40.4\%} 
   	&\color{mygreen}{$\uparrow$51.2\%} 
   	&\color{mygreen}{$\uparrow$28.3\%}
   	&\color{mygreen}{$\uparrow$2.5\%}
   	&\color{mygreen}{$\uparrow$18.8\%}
   	&\color{mygreen}{$\uparrow$16.6\%}
   	&\color{mygreen}{$\uparrow$16.4\%}
   	&\color{mygreen}{$\uparrow$19.8\%}
   	\\
   	
   	\hline \hline
   	\multirow{6}{*}{Average} 
   	&SAM 2~\cite{SAM2}
   	&ICLR 2025&Hiera-B+~\cite{Hiera}
   	& 0.9829& 0.1884& 0.1663& 0.1520 & 0.5245 &0.7136 &0.5455 &0.1421& 0.4231& 0.0184
   	
   	\\
   	&SegGPT~\cite{SegGPT}&ICCV 2023 & ViT-L~\cite{ViT} 
   	& 0.9818 & 0.1472 & 0.1332  &0.1300 &   0.5284  &0.6697  &0.5331  &0.1316  &0.4259& 0.0181
   	
   	\\
   	&DeepLabV3+~\cite{DeepLabV3+}
   	&ECCV 2018&ResNet-101~\cite{Resnet}
   	& 0.9870  & 0.3717  &0.3289 & 0.3145  &0.4297 & 0.8708 & 0.5012  &0.3707  &0.3296 & 0.0155
   	
   	\\
   	&SegFormer~\cite{SegFormer}
   	&NeurIPS 2021&MiT-B5~\cite{SegFormer}
   	& 0.9878& 0.3622 &0.3210  &0.3020  &0.4250  &0.8584 &0.5161 &0.3672 &0.3384 &0.0155

   	\\
   	&Spider~\cite{CDCU-Spider}
   	&ICML 2024&ConvNeXt-L~\cite{ConvNeXt}
   	& 0.9896& 0.4734 &0.4543& 0.4517 &0.5968 &0.9342& 0.5950  &0.4970&  0.2129 &0.0127
   	\\
   	&ZoomNeXt~\cite{ZoomNeXt}
   	&TPAMI 2024&ResNet-50~\cite{Resnet}
   	& 0.9888&  0.5134 & 0.4899 & 0.4810 &  0.6513&  0.9511 & 0.6290  & 0.5328&  0.1965&  0.0115
   	\\
   	\cline{2-14}
   	&MDCNet~\cite{MDCNet}
   	&CVPR 2024&ResNet-50~\cite{Resnet} 
   	& 0.9902  &0.5619  &0.5256 & 0.5291  &0.6840  & 0.9567 & 0.6662  &0.5811  &0.1738 & 0.0100 
   	\\
   	&MDCNeXt
   	&--&ResNet-50~\cite{Resnet} 
   	& \color{reda}\textbf{0.9912}  & \color{reda}\textbf{0.6232}&  \color{reda}\textbf{0.6225} &\color{reda}\textbf{0.6521} & \color{reda}\textbf{0.7699} & \color{reda}\textbf{0.9761} &  \color{reda}\textbf{0.7458}&  \color{reda}\textbf{0.6489}& \color{reda}\textbf{0.1494}&
   	\color{reda}\textbf{0.0087}
   	\\
   	
   	&    &  & 
   	& \color{mygreen}{$\uparrow$0.1\%} 
   	& \color{mygreen}{$\uparrow$10.9\%} 
   	& \color{mygreen}{$\uparrow$18.4\%} 
   	&\color{mygreen}{$\uparrow$23.2\%} 
   	&\color{mygreen}{$\uparrow$12.6\%}
   	&\color{mygreen}{$\uparrow$2.0\%}
   	&\color{mygreen}{$\uparrow$11.9\%}
   	&\color{mygreen}{$\uparrow$11.7\%}
   	&\color{mygreen}{$\uparrow$14.3\%}
   	&\color{mygreen}{$\uparrow$13.0\%}
   	\\
   	\bottomrule[2pt]
   \end{tabular}
   
	}
	\label{tab:comparison_segmentation}
\end{table*}

\begin{figure*}[!t]
\centering
\includegraphics[width=0.9\linewidth]{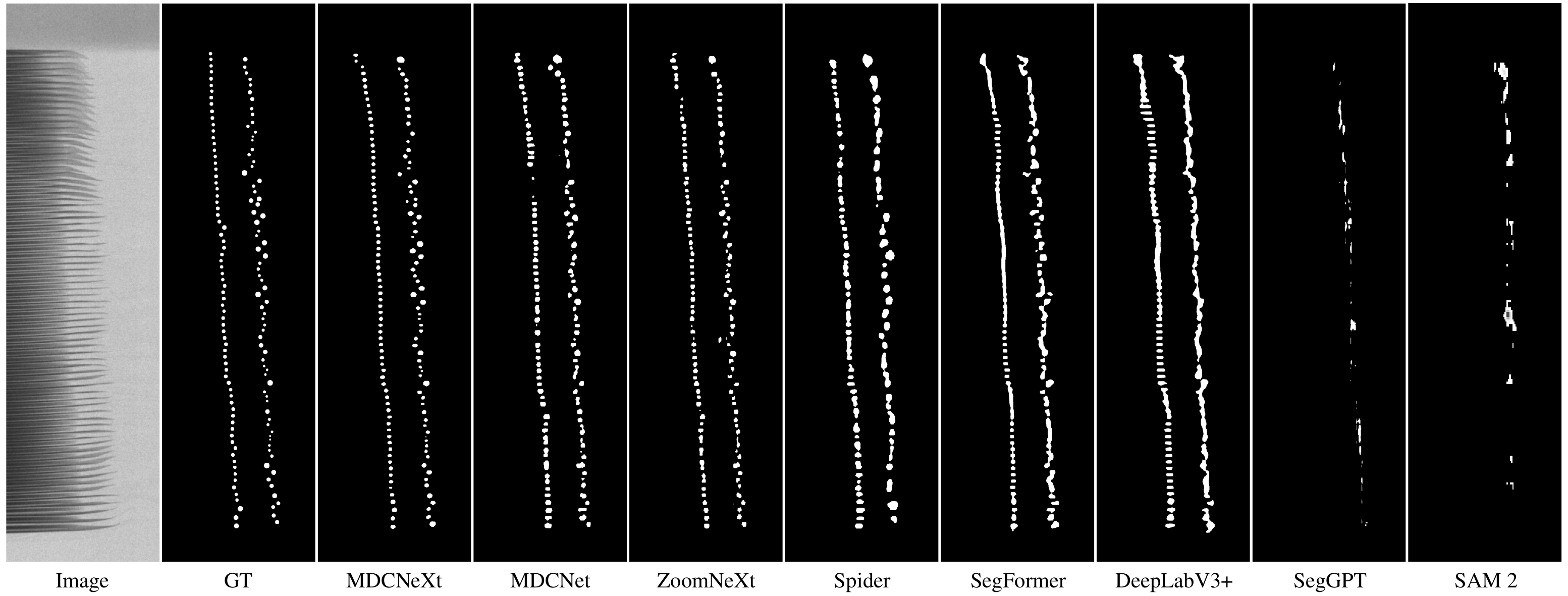}
\caption{Visual comparison with different image segmentation methods on binarized point maps for both anode and cathode.}
\label{fig:segmentation_compar}
\end{figure*}
\textbf{Qualitative Comparison.} 
Fig.~\ref{fig:Visual_results} presents visual predictions of MDCNeXt across diverse battery configurations. Tough sample is a high-resolution image with over 70 plate layers, where dense stacking, fine-scale structures, and boundary fluctuations make the task highly challenging. Difficult samples involve the large overhang, illumination interference, and tightly packed distributions. Regular samples present bifurcation interference, separator interference and pure plate structures. In all these scenarios, MDCNeXt achieves consistently precise and robust predictions. 
Fig.~\ref{fig:other_compar_mdcnext} shows visual comparisons with other methods. MDCNet produces prediction errors at the endpoints when facing bifurcation interference and shows adhesion issues in dense regions. Box-based detection methods (HS-FPN, CFINet, YOLOv10) fail under bifurcations, trays, and separators, often producing redundant boxes prediction. 
 DETR and RetinaNet are unable to overcome illumination interference, leading to many missed detections.
Counting methods provide rough density maps but cannot localize each plate individually. 
Corner detection methods will yield redundant point maps and can not distinguish plate endpoints from general intersection points even if we have performed edge extraction to pre-filter a large amount of background irrelevant to plate endpoints.

\subsection{Comparison with Image Segmentation Methods} 
\textbf{Quantitative Comparison.}      
To comprehensively evaluate MDCNeXt’s segmentation performance, we compare it with generalist models (SAM 2~\cite{SAM2}, SegGPT~\cite{SegGPT}), general segmentation frameworks (DeepLabV3+~\cite{DeepLabV3+}, SegFormer~\cite{SegFormer}), and context-dependent concept segmentation models (Spider~\cite{CDCU-Spider}, ZoomNeXt~\cite{ZoomNeXt}). As shown in Table~\ref{tab:comparison_segmentation}, ten widely adopted segmentation metrics are used for evaluation.
From the average scores, SAM 2 and SegGPT perform reasonably on $PA$ and $S_m$ due to the dominance of background regions and structure-aware properties, but underperform on other metrics. This suggests that these in-context learning-based models struggle to propagate prompt features into the main segmentation branch for fine-grained guidance, limiting their adaptability to tasks like PBD even with large-scale pretraining.
In contrast, Spider and ZoomNeXt benefit from foreground-background comparison and rich-scale fusion, outperforming general segmentation methods. 
Notably, MDCNeXt achieves the best results across all metrics. On the tough split, it surpasses MDCNet with improvements of 51.2\% in $F_\beta^w$, 18.8\% in $mIoU$, and 19.8\% in $\mathcal{M}$, showing the effectiveness of our prompt-filtered and density-aware reordering state space modules. Tab.~\ref{tab:efficiency_comparison} further presents the efficiency comparison. MDCNeXt still shows clear advantages, achieving a good balance between parameter size and computation.
{To show the advantages of MDCNeXt in adopting SSM, we conduct runtime and memory comparisons between MDCNeXt and strong Transformer-based baselines, with a particular focus on high-resolution inputs, as summarized in Tab.~\ref{tab:runtime_memory}.
Specifically, we report FPS and peak GPU memory usage at both $512\times512$ and $2048\times2048$ resolutions under a unified setting (batch size $=1$, single A100-40GB GPU). 
Transformer-based models such as SegFormer-B5, Spider (Swin-L), and SAM 2 (Hierarchical-B+)  experience rapidly increasing memory consumption as input resolution grows, due to the quadratic complexity of self-attention with respect to token number. 
At $2048\times2048$, these models encounter out-of-memory (OOM) errors, while MDCNeXt, leveraging a linear-complexity state space model, maintains stable memory usage and latency growth.
In fact, there are potential optimization strategies in real-world deployment such as model pruning, mixed-precision inference, and hardware-specific acceleration. They can be readily applied without modifying the core model design. These engineering-level optimizations are complementary to our method and will be explored in future work to further enhance real-time performance in resource-constrained industrial environments.
}

\begin{table}[!t]
\centering
\caption{Efficiency comparison of different segmentation methods in terms of parameter size and computational cost. The best two results are highlighted in {\color{reda} \textbf{red}} and {\color{mygreen} \textbf{green}}, respectively. The worst results are marked with \underline{underline}.}

   \resizebox{\linewidth}{!}{
    \setlength\tabcolsep{2pt}
    \renewcommand\arraystretch{1.2}
\begin{tabular}{r||c|c|c|c|c|c|c|c}
	\toprule[2pt]
	Metrics  & SAM 2 	& SegGPT & DeepLabV3+& SegFormer & Spider & ZoomNeXt  &MDCNet&MDCNeXt
	\\
	&~\cite{SAM2}& ~\cite{SegGPT} & ~\cite{DeepLabV3+}& ~\cite{SegFormer} & ~\cite{CDCU-Spider} & ~\cite{ZoomNeXt}  &~\cite{MDCNet}& (Ours)
	\\
	\hline
	Parameters (MB)$\downarrow$
	&216.40& 370.26& 58.94&84.75& \underline{394.21} &\color{reda} \textbf{28.46}& 49.54& \color{mygreen} \textbf{41.54} 
	\\
	GFLOPs$\downarrow$
	&128.40 &\underline{655.65} &121.54 &372.52 &171.42 &150.02  &\color{reda} \textbf{27.62} &\color{mygreen} \textbf{73.16} 
	\\
	\bottomrule[2pt]
\end{tabular}

	}
	\label{tab:efficiency_comparison}
\end{table}

\begin{table}[t]
\centering
	\caption{{Runtime and memory comparison between MDCNeXt and strong Transformer-based segmentation methods at different input resolutions.} 
}
   \resizebox{0.8\linewidth}{!}{
    \setlength\tabcolsep{2pt}
    \renewcommand\arraystretch{1.2}
  \begin{tabular}{c||cc|cc}
 \toprule[2pt]
\multirow{2}{*}{Method} 
& \multicolumn{2}{c|}{512$\times$512} 
& \multicolumn{2}{c}{2048$\times$2048} 
\\
\cline{2-3} \cline{4-5}
& FPS$\uparrow$ & Mem (GB)$\downarrow$
& FPS$\uparrow$ & Mem (GB)$\downarrow$ 
\\
	\hline
SegFormer-B5    
& 21 & 3.9
& --   & OOM \\

SAM 2 (Hierarchical-B+) 
& 31 & 6.8
& --   & OOM \\

Spider (Swin-L)
& 38 & 5.7
& --   & OOM \\
	\hline
{MDCNeXt} 
& \color{reda}\textbf{54} & \color{reda}\textbf{3.2}
&\color{reda}\textbf{6}  &\color{reda}\textbf{18.4} 
\\
\bottomrule[2pt]
\end{tabular}
	}
\label{tab:runtime_memory}
\end{table}

\textbf{Qualitative Comp
arison.}
Fig.~\ref{fig:segmentation_compar} shows the qualitative results on a tough sample. MDCNeXt outputs points with sizes and spacing aligned to local variations, showing high consistency with the ground truth. MDCNet and ZoomNeXt show point adhesions and discontinuities, while Spider suffers from merging multiple adjacent points into large responses. SegFormer and DeepLabV3+ fail to localize individual points, generating line-like outputs. SegGPT and SAM 2 barely capture plate structures, resulting in poor predictions. Overall, the visualization highlights the comprehensive capability of each model in handling scale variation, spatial precision, and feature discrimination.

\subsection{Cross-Type Generalization}
{As shown in Tab.~\ref{tab:comparision_each_attribute} and Tab.~\ref{tab:comparison_seg_each_attribute}, we compare MDCNeXt with representative top-performing detection and segmentation methods for the \textbf{battery-category–wise evaluation}. We can see that while absolute performance varies with structural complexity (e.g., plate density and interference severity), MDCNeXt consistently outperforms competing methods across all battery types, indicating stable cross-type generalization rather than reliance on type-specific biases.}

\begin{table}[t]
\centering
\caption{{Quantitative comparison of MDCNeXt with  top-performing tiny object detection methods for the \textbf{battery-category-wise evaluation}. $\uparrow$ and $\downarrow$ indicate that higher and lower values are better, respectively. The best results are highlighted in \color{reda}\textbf{red}.}}
   \resizebox{\linewidth}{!}{
    \setlength\tabcolsep{2pt}
    \renewcommand\arraystretch{1.2}
\begin{tabular}{c|r||c|c|c|c|c|c|c|c}
\toprule[2pt]
     Attribute&Method  & AN-MAE$\downarrow$ 	& CN-MAE$\downarrow$ & AN-ACC$\uparrow$ & CN-ACC$\uparrow$ & PN-ACC$\uparrow$ & AL-MAE$\downarrow$  &CL-MAE$\downarrow$&OH-MAE$\downarrow$
     \\
     				\hline

\multirow{4}{*}{\textbf{P}}  
&CFINet
 &1.5452
 &1.0542
 &0.7802
 &0.7520
 &0.7040
 &2.0554
 &2.0014
 &2.0471
 \\
&HS-FPN
&1.1075
&1.0021
&0.7856
&0.7740
&0.7165
&2.0771
&1.9576
&2.2045
 \\
&MDCNet
&0.2874
&0.1845
&0.9413
&0.9304
&0.8841
&1.3042
&1.3144
&1.3408
 \\
&MDCNeXt
&\color{reda}\textbf{0.2102}
&\color{reda}\textbf{0.1308}
&\color{reda}\textbf{0.9901}
&\color{reda}\textbf{0.9874}
&\color{reda}\textbf{0.9532}
&\color{reda}\textbf{1.0384}
&\color{reda}\textbf{1.1030}
&\color{reda}\textbf{1.0004}
 \\
 \hline
 \multirow{4}{*}{\textbf{T}}  
&CFINet
&3.5463
&2.8764
&0.3452
&0.3051
&0.2251
&8.0215
&8.8845
&7.0522
 \\
 &HS-FPN
 &2.7452
&2.0845
&0.4513
&0.4015
&0.2815
&6.0215
&5.7454
&5.5515 
 \\
 &MDCNet
&0.9123
&0.7521
&0.7851
&0.7626
&0.7011
&1.4512
&1.5612
&1.3020
 \\
&MDCNeXt
&\color{reda}\textbf{0.5315}
&\color{reda}\textbf{0.3015}
&\color{reda}\textbf{0.8765}
&\color{reda}\textbf{0.8345}
&\color{reda}\textbf{0.8086}
&\color{reda}\textbf{1.0125}
&\color{reda}\textbf{0.8451}
&\color{reda}\textbf{0.9815}
 \\
 \hline
  \multirow{4}{*}{\textbf{A}}  
 &CFINet
 &3.4521
&3.7102
&0.4521
&0.4102
&0.3025
&8.1252
&8.0415
&7.3521
 \\
 &HS-FPN
&1.4512
&1.4012
&0.5412
&0.4915
&0.4350
&8.9412
&8.0100
&7.8951 
 \\
&MDCNet
&0.4315
&0.4011
&0.8215
&0.7915
&0.7345
&4.3541
&4.8415
&3.7754
 \\
&MDCNeXt
&\color{reda}\textbf{0.3322}
&\color{reda}\textbf{0.3215}
&\color{reda}\textbf{0.9015}
&\color{reda}\textbf{0.8991}
&\color{reda}\textbf{0.8452}
&\color{reda}\textbf{1.4120}
&\color{reda}\textbf{1.2250}
&\color{reda}\textbf{1.1815}
 \\
 \hline
  \multirow{4}{*}{\textbf{II}}  
 &CFINet
&10.5415
&9.8485
&0.1642
&0.1854
&0.1036
&18.2596
&18.0200
&11.4546
 \\
 &HS-FPN
&9.4125
&9.0021
&0.1954
&0.2031
&0.1602
&20.5412
&18.0215
&7.8454 
 \\
 &MDCNet
&6.8462
&4.9785
&0.4215
&0.3874
&0.3321
&9.4152
&8.8452
&6.0754
 \\
&MDCNeXt
&\color{reda}\textbf{0.7451}
&\color{reda}\textbf{0.3536}
&\color{reda}\textbf{0.7475}
&\color{reda}\textbf{0.7109}
&\color{reda}\textbf{0.6102}
&\color{reda}\textbf{1.1021}
&\color{reda}\textbf{0.9426}
&\color{reda}\textbf{0.9502}
 \\
 \hline
  \multirow{4}{*}{\textbf{PI}}  
 &CFINet
&5.8941
&6.0315
&0.3152
&0.2842
&0.1887
&12.4512
&12.0845
&9.8785
 \\
 &HS-FPN
&4.3120
&4.8450
&0.3877
&0.3511
&0.2539
&10.8412
&12.0150
&10.3152 
 \\
 &MDCNet
&2.7451
&2.2341
&0.6084
&0.5702
&0.5041
&4.0215
&5.0415
&3.8845
 \\
&MDCNeXt
&\color{reda}\textbf{0.6512}
&\color{reda}\textbf{0.3451}
&\color{reda}\textbf{0.8310}
&\color{reda}\textbf{0.8022}
&\color{reda}\textbf{0.7151}
&\color{reda}\textbf{1.0215}
&\color{reda}\textbf{1.1220}
&\color{reda}\textbf{0.9025}
 \\
 \hline
  \multirow{4}{*}{\textbf{BI}}  
 &CFINet
 &1.9849
 &1.6051
 &0.6687
 &0.6846
 &0.6177
 &2.4151
 &2.3705
 &2.4002
 \\
&HS-FPN
&1.4751
&1.1348
&0.6975
&0.7105
&0.6564
&2.5012
&2.0454
&2.2414
 \\
&MDCNet
&0.3408
&0.2307
&0.8715
&0.8797
&0.8510
&1.4784
&1.3871
&1.3940
 \\
&MDCNeXt
 &\color{reda}\textbf{0.2286}
&\color{reda}\textbf{0.1308}
&\color{reda}\textbf{0.9254}
&\color{reda}\textbf{0.9502}
&\color{reda}\textbf{0.9087}
&\color{reda}\textbf{1.1741}
&\color{reda}\textbf{1.2075}
&\color{reda}\textbf{1.0820}
 \\
 \hline
  \multirow{4}{*}{\textbf{TRI}}  
&CFINet
 &2.0021
 &1.5036
 &0.6410
 &0.6254
 &0.6017
 &2.4947
 &2.4064
 &2.4545
 \\
 &HS-FPN
 &1.4035
&1.1807
&0.6998
&0.7208
&0.6702
&2.3811
&2.1506
&2.4217
 \\
&MDCNet
&0.3175
&0.2451
&0.8785
&0.8745
&0.8302
&1.3805
&1.4310
&1.4038
 \\
&MDCNeXt
 &\color{reda}\textbf{0.2206}
&\color{reda}\textbf{0.1329}
&\color{reda}\textbf{0.9210}
&\color{reda}\textbf{0.9308}
&\color{reda}\textbf{0.9020}
&\color{reda}\textbf{1.1402}
&\color{reda}\textbf{1.0221}
&\color{reda}\textbf{1.0754}
 \\
 \hline
  \multirow{4}{*}{\textbf{TAI}}  
 &CFINet
&3.4520
&3.7462
&0.4441
&0.3844
&0.2432
&9.0274
&8.4512
&7.0210
 \\
 &HS-FPN
 &1.9742
&2.0472
&0.5210
&0.4041
&0.3456
&8.0210
&8.4550
&7.8465 
 \\
 &MDCNet
&0.5435
&0.4841
&0.7385
&0.7231
&0.7072
&4.0214
&4.7462
&3.7542
 \\
&MDCNeXt
&\color{reda}\textbf{0.3847}
&\color{reda}\textbf{0.3715}
&\color{reda}\textbf{0.8476}
&\color{reda}\textbf{0.8215}
&\color{reda}\textbf{0.7765}
&\color{reda}\textbf{1.2141}
&\color{reda}\textbf{1.2021}
&\color{reda}\textbf{1.2012}
 \\
 \hline
  \multirow{4}{*}{\textbf{SI}}  
 &CFINet
 &2.7450
&2.8400
&0.5376
&0.4875
&0.3451
&9.9756
&9.4514
&7.8785
 \\
 &HS-FPN
 &1.0421
&1.1075
&0.6544
&0.5438
&0.4375
&9.4475
&9.8475
&9.0145 
 \\
 &MDCNet
 &0.4401
&0.4372
&0.8478
&0.8028
&0.7745
&4.8421
&4.4700
&4.8784
 \\
&MDCNeXt
 &\color{reda}\textbf{0.3107}
&\color{reda}\textbf{0.3002}
&\color{reda}\textbf{0.9374}
&\color{reda}\textbf{0.9187}
&\color{reda}\textbf{0.8830}
&\color{reda}\textbf{1.2877}
&\color{reda}\textbf{1.2980}
&\color{reda}\textbf{1.2002}
 \\
\bottomrule[2pt]
\end{tabular}
	}
\label{tab:comparision_each_attribute}
\end{table}
\begin{table}[t]
\centering
	\caption{{Quantitative comparison of MDCNeXt with  top-performing segmentation methods for the \textbf{battery-category-wise evaluation}. $\uparrow$ and $\downarrow$ indicate that higher and lower values are better, respectively. The best results are highlighted in \color{reda}\textbf{red}.}}
   \resizebox{\linewidth}{!}{
    \setlength\tabcolsep{2pt}
    \renewcommand\arraystretch{1.2}
\begin{tabular}{c|r||c|c|c|c|c|c|c|c|c|c}
\toprule[2pt]
     Attribute&Method  & $PA\uparrow$ 	& $F_{\beta}^{max}\uparrow$ & $F_{\beta}^{mean}\uparrow$ & $F_{\beta}^{\omega}\uparrow$ & $S_m\uparrow$ & $E_m\uparrow$  &$mIoU\uparrow$&$mDice\uparrow$  &$BER\downarrow$ &$\mathcal{M}\downarrow$ 
     \\
     				\hline

\multirow{4}{*}{\textbf{P}}  
&Spider
&0.9888
&0.6425
&0.5970
&0.6212
&0.7836
&0.9425
&0.6820
&0.6278
&0.1422
&0.0097
 \\
 &ZoomNeXt
&0.9902
&0.6507
&0.5988
&0.6380
&0.7971
&0.9635
&0.7135
&0.6685
&0.1270
&0.0089  
 \\
 &MDCNet
&0.9914
&0.6920
&0.6754
&0.7154
&0.8036
&0.9542
&0.7652
&0.7441
&0.1095
&0.0079
 \\
&MDCNeXt
&\color{reda}\textbf{0.9942}
&\color{reda}\textbf{0.7351}
&\color{reda}\textbf{0.7152}
&\color{reda}\textbf{0.7644} 
&\color{reda}\textbf{0.8511}
&\color{reda}\textbf{0.9850}
&\color{reda}\textbf{0.8146}
&\color{reda}\textbf{0.7735}
&\color{reda}\textbf{0.1038}
&\color{reda}\textbf{0.0070}
 \\
 \hline
 \multirow{4}{*}{\textbf{T}}  
&Spider
 &0.9902
&0.4021
&0.3644
&0.3452
&0.4842
&0.9309
&0.5451
&0.4345
&0.2302
&0.0137
 \\
 &ZoomNeXt
 &0.9887
 &0.4302
 &0.3846
 &0.3745
 &0.5021
 &0.9454
 &0.5897
 &0.4941
 &0.2152
 &0.0126
 \\
 &MDCNet
&0.9898
&0.4852
&0.4414
&0.3960
&0.5845
&0.9509
&0.6312
&0.5274
&0.2078
&0.0115
 \\
&MDCNeXt
&\color{reda}\textbf{0.9908}
&\color{reda}\textbf{0.5764}
&\color{reda}\textbf{0.5694}
&\color{reda}\textbf{0.5846}
&\color{reda}\textbf{0.7541}
&\color{reda}\textbf{0.9774}
&\color{reda}\textbf{0.7456}
&\color{reda}\textbf{0.6021}
&\color{reda}\textbf{0.1802}
&\color{reda}\textbf{0.0094}
 \\
 \hline
  \multirow{4}{*}{\textbf{A}}  
&Spider
 &0.9902
&0.4845
&0.5021
&0.4845
&0.5914
&0.9444
&0.6215
&0.5312
&0.2021
&0.0115
 \\
 &ZoomNeXt
 &0.9892
&0.5001
&0.5092
&0.5045
&0.6951
&0.9514
&0.6378
&0.5451
&0.1945
&0.0102
 \\
 &MDCNet
 &0.9912
&0.5854
&0.5642
&0.5394
&0.7251
&0.9590
&0.6951
&0.5841
&0.1692
&0.0091
 \\
&MDCNeXt
 &\color{reda}\textbf{0.9918}
&\color{reda}\textbf{0.6451}
&\color{reda}\textbf{0.6366}
&\color{reda}\textbf{0.6785}
&\color{reda}\textbf{0.7760}
&\color{reda}\textbf{0.9764}
&\color{reda}\textbf{0.7674}
&\color{reda}\textbf{0.6875}
&\color{reda}\textbf{0.1402}
&\color{reda}\textbf{0.0082}
 \\
 \hline
  \multirow{4}{*}{\textbf{II}}  
 &Spider
&0.9800
&0.3541
&0.3021
&0.3002
&0.4214
&0.9252
&0.4956
&0.4115
&0.2536
&0.0152
 \\
 &ZoomNeXt
 &0.9802
 &0.3784
 &0.3412
 &0.3032
 &0.5021
 &0.9377
 &0.5021
 &0.4411
 &0.2487
 &0.0149
 \\
&MDCNet
&0.9820
&0.4210
&0.3452
&0.3215
&0.5145
&0.9402
&0.5421
&0.4675
&0.2402
&0.0154
 \\
&MDCNeXt
&\color{reda}\textbf{0.9900}
&\color{reda}\textbf{0.5451}
&\color{reda}\textbf{0.5530}
&\color{reda}\textbf{0.5652}
&\color{reda}\textbf{0.7021}
&\color{reda}\textbf{0.9702}
&\color{reda}\textbf{0.7054}
&\color{reda}\textbf{0.5641}
&\color{reda}\textbf{0.1887}
&\color{reda}\textbf{0.0102}
 \\
 \hline
  \multirow{4}{*}{\textbf{PI}}  
&Spider
&0.9899
&0.3985
&0.3451
&0.3394
&0.4754
&0.9305
&0.5341
&0.4154
&0.2398
&0.0140
 \\
 &ZoomNeXt
 &0.9890
 &0.4451
 &0.3802
 &0.3678
 &0.5202
 &0.9484
 &0.5854
 &0.4895
 &0.2187
 &0.0127
 \\
 &MDCNet
&0.9893
&0.4821
&0.4312
&0.3885
&0.5774
&0.9525
&0.6157
&0.5174
&0.2120
&0.0114
 \\
&MDCNeXt
&\color{reda}\textbf{0.9908}
&\color{reda}\textbf{0.5752}
&\color{reda}\textbf{0.5694}
&\color{reda}\textbf{0.5865}
&\color{reda}\textbf{0.7541}
&\color{reda}\textbf{0.9762}
&\color{reda}\textbf{0.7355}
&\color{reda}\textbf{0.5954}
&\color{reda}\textbf{0.1808}
&\color{reda}\textbf{0.0095}
 \\
 \hline
  \multirow{4}{*}{\textbf{BI}}  
&Spider
&0.9802
&0.6075
&0.5844
&0.6202
&0.7542
&0.9451
&0.6745
&0.6021
&0.1502
&0.0099
 \\
 &ZoomNeXt
&0.9843
&0.6085
&0.5974
&0.6225
&0.7541
&0.9544
&0.6840
&0.6451
&0.1488
&0.0097  
 \\
&MDCNet
&0.9902
&0.6645
&0.6541
&0.6974
&0.7901
&0.9644
&0.7275
&0.7064
&0.1146
&0.0084
 \\
&MDCNeXt
&\color{reda}\textbf{0.9917}
&\color{reda}\textbf{0.6902}
&\color{reda}\textbf{0.6921}
&\color{reda}\textbf{0.7245} 
&\color{reda}\textbf{0.8004}
&\color{reda}\textbf{0.9754}
&\color{reda}\textbf{0.7754}
&\color{reda}\textbf{0.7284}
&\color{reda}\textbf{0.1126}
&\color{reda}\textbf{0.0078}
 \\
 \hline
  \multirow{4}{*}{\textbf{TRI}}  
&Spider
&0.9785
&0.6354
&0.5984
&0.6025
&0.7542
&0.9502
&0.6754
&0.6011
&0.1454
&0.0098
 \\
&ZoomNeXt
&0.9802
&0.6275
&0.6096
&0.6275
&0.7645
&0.9602
&0.6845
&0.6125
&0.1384
&0.0094  
 \\
&MDCNet
&0.9846
&0.6674
&0.6641
&0.6954
&0.7850
&0.9645
&0.7215
&0.7113
&0.1202
&0.0084
 \\
&MDCNeXt
&\color{reda}\textbf{0.9902}
&\color{reda}\textbf{0.6911}
&\color{reda}\textbf{0.6845}
&\color{reda}\textbf{0.7215} 
&\color{reda}\textbf{0.8052}
&\color{reda}\textbf{0.9791}
&\color{reda}\textbf{0.7754}
&\color{reda}\textbf{0.7385}
&\color{reda}\textbf{0.1144}
&\color{reda}\textbf{0.0079}
  \\
 \hline
  \multirow{4}{*}{\textbf{TAI}}  
&Spider
 &0.9897
&0.4475
&0.4456
&0.4354
&0.6125
&0.9289
&0.5875
&0.4914
&0.2398
&0.0134
 \\
 &ZoomNeXt
&0.9832
&0.4987
&0.5003
&0.4905
&0.6645
&0.9495
&0.6021
&0.5205
&0.2231
&0.0121
 \\
 &MDCNet
&0.9902
&0.5461
&0.5324
&0.5156
&0.6945
&0.9503
&0.6541
&0.5643
&0.1724
&0.0092
 \\
&MDCNeXt
&\color{reda}\textbf{0.9909}
&\color{reda}\textbf{0.6013}
&\color{reda}\textbf{0.6052}
&\color{reda}\textbf{0.6516}
&\color{reda}\textbf{0.7701} 
&\color{reda}\textbf{0.9703}
&\color{reda}\textbf{0.7403}
&\color{reda}\textbf{0.6515}
&\color{reda}\textbf{0.1452}
&\color{reda}\textbf{0.0085}
 \\
 \hline
  \multirow{4}{*}{\textbf{SI}}  
&Spider
&0.9910
&0.4865
&0.4715
&0.4554
&0.6345
&0.9384
&0.6265
&0.5145
&0.2024
&0.0119
 \\
 &ZoomNeXt
 &0.9891
&0.5014
&0.5002
&0.4892
&0.6941
&0.9584
&0.6453
&0.5402
&0.2132
&0.0107
 \\
 &MDCNet
 &0.9911
&0.5668
&0.5542
&0.5402
&0.7210
&0.9565
&0.6785
&0.5964
&0.1641
&0.0091
 \\
&MDCNeXt
&\color{reda}\textbf{0.9915}
&\color{reda}\textbf{0.6452}
&\color{reda}\textbf{0.6402}
&\color{reda}\textbf{0.6865}
&\color{reda}\textbf{0.7846} 
&\color{reda}\textbf{0.9768}
&\color{reda}\textbf{0.7645}
&\color{reda}\textbf{0.6845}
&\color{reda}\textbf{0.1404}
&\color{reda}\textbf{0.0082}
 \\
\bottomrule[2pt]
\end{tabular}

	}
\label{tab:comparison_seg_each_attribute}
\end{table}

\begin{table}[!t]
\centering
	\caption{{Ablation experiments of each component in the MDCNeXt.}}
   \resizebox{\linewidth}{!}{
    \setlength\tabcolsep{2pt}
    \renewcommand\arraystretch{1.2}
    \begin{tabular}{c|r||c|c|c|c|c|c|c|c|c}
    	\toprule[2pt]
    	Dataset&Method  &{Para.$\downarrow$}& AN-MAE$\downarrow$ 	& CN-MAE$\downarrow$ & AN-ACC$\uparrow$ & CN-ACC$\uparrow$ & PN-ACC$\uparrow$ & AL-MAE$\downarrow$  &CL-MAE$\downarrow$&OH-MAE$\downarrow$
    	\\
    	\hline

    	\multirow{5}{*}{Regular}  
    	&Baseline
        &25.42
    	&2.3821 &	2.0212 	&	0.7023 &0.7320 &0.6792 &2.0321 &2.1623 &2.4392 
    	\\
    	&+ CP
        &25.97
    	&1.5320 &	1.5012 	&	0.7911 &0.7932 &0.7509 &2.1023 &2.1123 &2.4086  
    	\\
    	&+ LP
        &27.03
    	&1.1021 &	1.0023 	&	0.8212 &0.8412 &0.8032 &1.5320 &1.4782 &1.5633 
    	\\
    	&+ Prompt Stream 
        &31.84
    	&0.5077 &	0.5632 	&	0.8744 &0.8664 &0.8398 &1.5402 &1.4402 &1.5222 
    	\\
    	&+ PFSSM
        &38.02
    	&0.2288 &	0.1794 	&	0.9132 &   0.9230 &    0.8746 &1.2722 &1.3021 &1.0432
    	\\
    	&+ DRSSM
        &41.54
    	&0.2175 &	0.1320 	&	0.9379 &	0.9456 &	0.9165 &	1.1537 &	1.2234 &	1.0102 
    	\\
    	\hline
    	\multirow{5}{*}{Difficult}  
    	&Baseline
        &25.42
    	&2.7982 &	2.6200 	&	0.5021 &0.4922 &0.4132 &7.1211 &7.1069 &5.3988 
    	\\
    	&+ CP
            &25.97
    	&1.9921 &	1.8902 	&	0.5794 &0.5633 &0.5090 &7.0232 &7.2109 &5.0231  
    	\\
    	&+ LP
            &27.03
    	&1.7032 &	1.5022 	&	0.6211 &0.6420 &0.5543 &5.0330 &5.1211 &3.0232 
    	\\
    	&+ Prompt Stream 
            &31.84
    	&0.8973 &	0.9012 	&	0.7342 &0.7212 &0.6532 &4.4547 &4.7034 &3.3203 
    	\\
    	&+ PFSSM
            &38.02
    	&0.4299 &	0.4023 	&	0.8302 &	0.8023 &0.7232 &2.2320 &2.5460 &1.7721  
    	\\
    	&+ DRSSM
            &41.54
    	&0.3515 &	0.3428 	&	0.8973 &	0.8552 &	0.7983 &1.3702 &1.2953 &1.1603 
    	\\
    	\hline
    	\multirow{5}{*}{Tough}  
    	&Baseline
            &25.42
    	&8.2110 &	7.5456&	0.3105 &0.3526 &0.2075 &	13.4512 &	12.0812 &	6.4516 
    	\\
    	&+ CP
            &25.97
    	&6.7008 &	6.4545&	0.3750 &0.4100 &0.2575 &	12.2545 &	12.9697 &	6.8510 
    	
    	\\
    	&+ LP
            &27.03
    	&6.3907 &	6.0762 	&0.4178 &0.4405 &0.3028 &	10.2010 & 9.1250 &6.3444 
    	\\
    	&+ Prompt Stream
            &31.84
    	&5.3210 &	4.5456 	&	0.5100 &	0.4846 &0.3912&9.1425 &	7.9450 &	5.9450 
    	\\
    	&+ PFSSM
            &38.02
    	&2.1021 &	1.5012 	&	0.6502 &	0.6271 &	0.5299 &3.4544 &2.0877 &	1.8946
    	\\
    	&+ DRSSM
        &41.54
    	&0.7873 &	0.3781 	&	0.7873 &	0.7341 &	0.6263 &1.2145 &	0.9826 &	0.9979 
    	\\
    	\hline \hline
    	\multirow{5}{*}{Average}   &Baseline
            &25.42
    	&4.5233 &4.1331 &0.4888 &0.5067& 0.4121 &7.9534 &7.5175& 4.9931
    	\\
    	&+ CP
            &25.97
    	&3.4675& 3.3350 & 0.5647& 0.5706 &0.4862 &7.5269 &7.8474 &4.9686 
    	
    	\\
    	&+ LP
            &27.03
    	&3.1351 &2.9218& 0.6038 &0.6251 &0.5333& 5.8809& 5.5384& 3.7715
    	\\
    	&+ Prompt Stream
            &31.84
    	&2.2944 &2.0478 &0.6944 &0.6785 &0.6126 &5.2910  &4.9604 &3.7458
    	\\
    	&+ PFSSM
            &38.02
    	&0.9442 &0.7169 &0.7906& 0.7741 &0.6968 &2.3986& 2.0706& 1.6259

    	\\
    	&+ DRSSM
            &41.54
    	&0.4645& 0.3005& 0.8705 &0.8375& 0.7705 &1.2617& 1.1709& 1.0667
    	\\
    	\bottomrule[2pt]
    \end{tabular}
    
	}
	\label{tab:ablation_study_modules}
\end{table}

\begin{table}[!t]
\centering
	\caption{{Ablation experiments of state space modeling, prompt filter and reordering operation in PFSSM and DRSSM. M1: w/o SSM in PFSSM. M2: w/o prompt filter in PFSSM. M3: fixed ordering operation in DRSSM. M4: random ordering operation in DRSSM. M5: replace DRSSM with multiple convolution layers.}}
   \resizebox{\linewidth}{!}{
    \setlength\tabcolsep{2pt}
    \renewcommand\arraystretch{1.2}
 \begin{tabular}{c|r||c|c|c|c|c|c|c|c|c}
 	\toprule[2pt]
 	Dataset&Method  & {Para.$\downarrow$}&AN-MAE$\downarrow$ 	& CN-MAE$\downarrow$ & AN-ACC$\uparrow$ & CN-ACC$\uparrow$ & PN-ACC$\uparrow$ & AL-MAE$\downarrow$  &CL-MAE$\downarrow$&OH-MAE$\downarrow$
 	\\
 	\hline

 	\multirow{7}{*}{Regular}  
 	&Baseline
    &25.42
 	&2.3821 &	2.0212 	&	0.7023 &0.7320 &0.6792 &2.0321 &2.1623 &2.4392 
 	\\
 	&MDCNeXt
       &41.54
 	&0.2175 &	0.1320 	&	0.9379 &	0.9456 &	0.9165 &	1.1537 &	1.2234 &	1.0102 
 	
 	\\
 	&M1
      &36.53
 	&0.2475 &	0.1812 	&	0.9109 &   0.9120 &    0.8956 & 1.2032 &1.2445 &1.0138
 	\\
 	&M2
     &39.49
 	&0.2511 &	0.1942 	&	0.9002 &   0.9020 &    0.8848 & 1.2205 &1.3214 &1.0836

 	\\
 	&M3
          &41.54
 	&0.2298 &	0.1777 	&	0.9103 &   0.9204 &    0.8702 &1.2790 &1.3056 &1.1890
 	
 	\\
     &M4
           &41.54
 &0.2674 &	0.1998 	&	0.8902 &   0.8984 &    0.8510 & 1.3745 &1.3954 &1.2514 
 \\
 	&M5
       &42.02
 	&0.2260 &	0.1780 	&	0.9121 &   0.9205 &    0.8764 &1.2571 &1.3126 &1.0733
 	
 	\\
 	\hline
 	\multirow{7}{*}{Difficult}  &Baseline
        &25.42
 	&2.7982 &	2.6200 	&	0.5021 &0.4922 &0.4132 &7.1211 &7.1069 &5.3988 
 	\\
 	&MDCNeXt
        &41.54
 	&0.3515 &	0.3428 	&	0.8973 &	0.8552 &	0.7983 &1.3702 &1.2953 &1.1603  
 	
 	\\
 	&M1
        &36.53
 	&0.4896 &	0.4288 	&	0.8025 &	0.8123 &	0.7436 &	1.7325 &	1.3646 &1.2754 
 	\\
 	&M2
        &39.49
 	&0.5346 &	0.5794 	&	0.8213 &	0.7787 &	0.7097 &	2.0252 &	2.0711 &1.4546

 	\\
 	&M3
        &41.54
 	&0.3954 &	0.3841 	&	0.8371 &	0.8021 &0.7152 &2.0984 &2.0715 &1.4635
 	\\
&M4
    &41.54
  &0.5074 &	0.5414 	&	0.7754 &	0.7456 &0.6477 &3.8488 &3.9241 &2.4302
 \\
 	&M5
     &42.02
 	&0.3947 &	0.3754 	&	0.8507 &	0.8211 &0.7325 &2.0741 &2.2150 &1.2513
 	
 	\\
 	\hline
 	\multirow{7}{*}{Tough}   &Baseline
        &25.42
 	&8.2110 &	7.5456&	0.3105 &0.3526 &0.2075 &	13.4512 &	12.0812 &	6.4516  
 	\\
 	&MDCNeXt
        &41.54
 	&0.7873 &	0.3781 	&	0.7873 &	0.7341 &	0.6263 &1.2145 &	0.9826 &	0.9979 
 	
 	\\
 	&M1
        &36.53
 	&1.8411 &	1.0215 	&	0.7092 &	0.6630 &	0.5416 &2.0211 &	1.4516 &	1.0321 
 	\\
 	&M2
        &39.49
 	&2.1641 &	1.5123 	&	0.6315 &	0.6406 &	0.5219 &2.2154 &	1.5145&	1.2541

 	\\
 	&M3
        &41.54
 	&2.1765 &	1.7121 	&	0.6741 &	0.6451 &	0.5468 &3.0976 &	2.1545 &	1.0812 
 	\\
  &M4
      &41.54
 &4.3084 &	3.9786 	&	0.5820 &	0.5079 &	0.4384 &5.2154 &4.3474 &	3.9021
  \\
 	&M5
        &42.02
 	&2.0084 &	1.5401 	&	0.6513 &	0.6289 &	0.5316 &3.0790 &2.1450 &	1.2015
 	\\
 	\hline \hline
 	\multirow{7}{*}{Average}    &Baseline
 	    &25.42
    &4.5233  &4.1331 & 0.4888  &0.5067 & 0.4121  &7.9534 & 7.5175  &4.9931
 	\\
 	&MDCNeXt
        &41.54
 	&0.4645& 0.3005& 0.8705 &0.8375 &0.7705& 1.2617& 1.1709 &1.0667
 	
 	\\
 	&M1
        &36.53
 	&0.8847 &0.5657 &0.7988 &0.7874 &0.7144 &1.6939 &1.3631 &1.1257
 	
 	\\
 	&M2
        &39.49
 	&1.0132 &0.7960 & 0.7774& 0.7637& 0.6912& 1.8824& 1.6896& 1.2912

 	\\
 	&M3
        &41.54
 	&0.9557 &0.7805 &0.8008 &0.7794 &0.6981& 2.2256 &1.9024 &1.2634
 	\\
     &M4
         &41.54
 &1.7322 & 1.6169  &0.7395  &0.7045 & 0.6292 & 3.6743  &3.4162 & 2.6249  
 \\
 	&M5
        &42.02
 	&0.8975 &0.7188 &0.7990  &0.7816 &0.7015 &2.2039& 1.9589& 1.1886
 	\\
 	\bottomrule[2pt]
 \end{tabular}
 
	}
	\label{tab:ablation_study_SSM}
\end{table}

\subsection{Ablation Study}
\textbf{Each Component.}
We analyze the contribution of each component in Tab.~\ref{tab:ablation_study_modules}. 
Our baseline is a point segmentation branch built on FPN~\cite{FPN} with a ResNet-50 backbone. 
First, we can see that the baseline has been able to outperform most competitors except HS-FPN, validating the suitability of point-based segmentation for the PBD task. 
Next, we add the counting predictor (CP) and line predictor (LP) to further refine the point segmentation at both high-level and low-level features. The former improves the accuracy of plate counting, while the latter helps adjust the positioning errors. Thus, multi-dimensional features complement each other and improve overall performance.
Then, we incorporate the prompt stream to guide the current image stream with pure plate features through simple convolution fusion. This addition provides consistent improvements across all splits under different metrics.
Finally, we introduce the state-space modeling into the architecture. PFSSM is integrated into the multi-level feature fusion between current and prompt streams. From the average results, we observe that ``+ PFSSM'' shows consistent improvements compared to ``+ Prompt Stream'' with the gain of 14\% and 57\% in terms of PN-ACC and OH-MAE. DRSSM is added at the tail of the framework to further refine the point maps. On the tough split, PN-ACC increases from 0.5299 to 0.6263, and AL/CL-MAE is reduced by more than 50\%. 

\textbf{Effectiveness of State Space Modeling, Prompt Filter and Reordering Operation.}
As shown in Tab.~\ref{tab:ablation_study_SSM}, we evaluate four variants of MDCNeXt. 
First, removing the state space modeling in PFSSM (M1) leads to significant performance degradation. Compared to the full MDCNeXt, the average AN-MAE and CN-MAE worsen from 0.4645 and 0.3005 to 0.8847 and 0.5657, while AN-ACC and CN-ACC drop from 0.8705 and 0.8375 to 0.7874 and 0.7144, respectively. These results confirm the critical role of SSM in modeling contextual relationships between the current and prompt streams.
Next, removing the prompt filter in PFSSM (M2) leads to a notable performance decline. The average PN-ACC drops from 0.7705 to 0.7062, while OH-MAE deteriorates from 1.0667 to 1.2912. These results indicate that the prompt filter is essential for refining the prompt signal and suppressing irrelevant interference. 
{Then, we conduct ablation studies comparing fixed ordering, random ordering, and density-aware reordering. Fixed ordering (M3) follows the default raster-scan flattening of the 2D feature map, while random ordering (M4) applies a random permutation over spatial tokens before state space modeling. 
We can see that density-aware reordering (MDCNeXt) consistently yields clear performance improvements, particularly on the tough split where plate density and structural ambiguity are most pronounced. 
Compared to density-aware reordering, fixed ordering already leads to noticeable performance degradation, as adjacent states often correspond to different plate instances or background regions, causing semantic interference during long-range state propagation.
Random ordering further exacerbates this issue and results in the severe performance drop. By completely destroying both spatial and semantic continuity, random ordering forces the state transitions to oscillate between unrelated structures, significantly impairing stable dependency modeling in dense and cluttered regions.
These results confirm that the observed gains do not stem from reordering itself, but specifically from aligning state transitions with physically meaningful plate structure and density.
Finally,  we replace DRSSM with multiple convolutional layers (M5) and obtain the same scale of parameters. We can see that M5 leads to consistent performance drops. Thus, simply stacking more convolutional layers cannot match the performance of the proposed reordering-based refinement. Moreover, the gap between M3/M4 and M5 suggests that directly applying SSM without the reordering mechanism in the final refinement stage loses its density-aware advantage, even underperforming vanilla convolution.}
 
\textbf{Prompt Attribute Diversity.}
In Tab.~\ref{tab:ablation_study_prompt_stable}, we evaluate the performance of the model using five different groups of \textbf{P} attribute images as the prompt input. 
It can be seen that MDCNeXt shows strong stability and generalization without dependence on a specific prompt. 
Thus, the strategy of random selection during training indeed makes MDCNeXt robust against different prompts when testing. 
{To further demonstrate the effectiveness of different prompts, we additionally analyze prompt attribute diversity. Specifically, we compare MDCNeXt models trained with all the nine different prompt attributes, and analyze their \textbf{performance, stability and convergence behavior}. From the results in Tab.~\ref{tab:ablation_study_attribute}, we can observe the following phenomena: 
1) From the performance perspective, 
pure plate prompts achieve the best overall performance, while other prompt attributes lead to varying degrees of performance degradation. 
Specifically, \textbf{BI}, \textbf{TRI} and \textbf{SI} exhibit relatively minor drops, whereas attributes involving large-area interference, such as \textbf{T}, \textbf{II} and \textbf{PI}, result in more pronounced degradation. On the one hand, these results highlight the importance of prompt selection, as an appropriate prompt can further enhance overall performance. On the other hand, they indicate that MDCNeXt does not rely on overly strict prompt conditions.  Even when using other non-pure attributes as prompts, our MDCNeXt still consistently maintains a clear performance advantage over competing methods, demonstrating the robustness and superiority of the overall framework.  
2) From the stability perspective  (see the ``Std.'' metric), pure plate prompts exhibit zero standard deviation, whereas other attributes show mild fluctuations due to the coexistence of multiple interfering factors. 
The prompt images in these experiments originate from different manufacturers and include moderate variations in exposure and imaging conditions. Despite this diversity, the filtering mechanism remains stable and does not break down during inference, indicating low sensitivity to prompt choice and robustness to realistic prompt quality variations.
3) From an optimization perspective, the prompt encoder is trained jointly with the main network, and the randomized prompts act as a form of structured regularization rather than noise injection. In practice, we observe stable convergence without oscillation or delayed training, indicating that prompt variability does not hinder optimization.}

\begin{table}[!t]
\centering
	\caption{Stability evaluation of MDCNeXt with randomly selected prompt inputs.}
   \resizebox{\linewidth}{!}{
    \setlength\tabcolsep{2pt}
    \renewcommand\arraystretch{1.2}
    \begin{tabular}{c||c|c|c|c|c|c|c|c}
    	\toprule[2pt]
    	No.  & AN-MAE$\downarrow$ 	& CN-MAE$\downarrow$ & AN-ACC$\uparrow$ & CN-ACC$\uparrow$ & PN-ACC$\uparrow$ & AL-MAE$\downarrow$  &CL-MAE$\downarrow$&OH-MAE$\downarrow$
    	\\
    	\hline

    	1
    	&0.4645& 0.3005& 0.8705 &0.8375& 0.7705 &1.2617& 1.1709& 1.0667 
    	\\
    	2
    	&0.4645& 0.3005& 0.8705 &0.8375& 0.7705 &1.2617& 1.1709& 1.0667 
    	
    	\\
    	3
    	&0.4645& 0.3005& 0.8705 &0.8375& 0.7705 &1.2617& 1.1709& 1.0667  
    	
    	\\
    	4
    	&0.4645& 0.3005& 0.8705 &0.8375& 0.7705 &1.2617& 1.1709& 1.0667 
    	\\
    	5
    	&0.4645& 0.3005& 0.8705 &0.8375& 0.7705 &1.2617& 1.1709& 1.0667 
    	\\
    	\bottomrule[2pt]
    \end{tabular}
    
	}
	\label{tab:ablation_study_prompt_stable}
\end{table}
\begin{table}[t]
\centering
	\caption{{Quantitative comparison of different prompt attributes.
Each row reports a model trained with a specific prompt attribute. Results are averaged over five inference runs with randomly sampled prompts and across the three test set splits. ``Std.'' denotes the standard deviation across the five runs, reflecting inference stability, while ``Epochs'' denotes the training convergence epoch range.}}
   \resizebox{\linewidth}{!}{
    \setlength\tabcolsep{2pt}
    \renewcommand\arraystretch{1.2}
\begin{tabular}{c||c|c|c|c|c|c|c|c|c|c}
 \toprule[2pt]
     Prompt Attribute  & AN-MAE$\downarrow$ 	& CN-MAE$\downarrow$ & AN-ACC$\uparrow$ & CN-ACC$\uparrow$ & PN-ACC$\uparrow$ & AL-MAE$\downarrow$  &CL-MAE$\downarrow$&OH-MAE$\downarrow$&  Std.$\downarrow$ & Epochs$\downarrow$
     \\
     				\hline
 \rowcolor[RGB]{235,235,250}
\textbf{P}
 &0.4645& 0.3005& 0.8705 &0.8375& 0.7705 &1.2617& 1.1709& 1.0667 & 0.0000 & 138-142 
 \\
\textbf{T}
 &0.5742& 0.4214& 0.8215 &0.7945& 0.7315 &1.4765& 1.6541& 1.2514 & 0.0055 & 138-142 
 \\
\textbf{A}
 &0.5241& 0.3765&  0.8317 &0.8025&0.7544 &1.4144& 1.2784& 1.2477 & 0.0038
& 138-142 
 
 \\
\textbf{II}
 &0.6487& 0.5715& 0.8117 &0.7879& 0.7214 &1.3874& 1.2476& 1.2020 & 0.0000 & 138-142 
 \\
\textbf{PI}
 &0.6754& 0.5902& 0.8028 &0.7802& 0.7114 &1.3421& 1.2845& 1.0929 & 0.0065 & 138-142 
  \\
\textbf{BI}
 &0.4876& 0.3154& 0.8587 &0.8274& 0.7657 &1.2874& 1.2541& 1.1745 & 0.0024 & 138-142
 \\
\textbf{TRI}
 &0.4899& 0.3102& 0.8579 &0.8285& 0.7697 &1.2803& 1.1948& 1.1145 & 0.0035 & 138-142 
 \\
\textbf{TAI}
 &0.5484& 0.4215& 0.8215 &0.7877& 0.7215 &1.3021& 1.1974& 1.0995 & 0.0000 & 138-142 
 \\
\textbf{SI}
 &0.4984& 0.3397& 0.8441 &0.8238& 0.7684 &1.2821& 1.2017& 1.0954 & 0.0054 & 138-142 
 \\
\bottomrule[2pt]
\end{tabular}

	}
\label{tab:ablation_study_attribute}
\end{table}

\textbf{Label Generation Strategies.} 
For point segmentation, we conduct a series of experiments about label generation strategies as discussed in Sec.~\ref{sec:generation_mask}.
As shown in Tab.~\ref{tab:ablation_study_label_strategy}, 
the model trained under the Ada-0.3 performs best in terms of number accuracy and localization error among all the six settings. The distance-adaptive  strategy consistently outperforms the constant strategy.

\begin{table}[!t]
\centering
\caption{Quantitative comparison of different label generation strategies and settings for the point segmentation branch. 
Const-1/3/5 refers to masks with radius of 1/3/5 pixels. 
The results are averaged across the three test set splits.}
   \resizebox{\linewidth}{!}{
    \setlength\tabcolsep{2pt}
    \renewcommand\arraystretch{1.2}
\begin{tabular}{c||c|c|c|c|c|c|c|c}
	\toprule[2pt]
	Method  & AN-MAE$\downarrow$ 	& CN-MAE$\downarrow$ & AN-ACC$\uparrow$ & CN-ACC$\uparrow$ & PN-ACC$\uparrow$ & AL-MAE$\downarrow$  &CL-MAE$\downarrow$&OH-MAE$\downarrow$
	\\
	\hline
	
	Const-1
	&6.8215& 7.3420& 0.3215 &0.3334& 0.2014 &14.2512& 11.0842& 6.0204 
	\\
	Const-3
	&3.1415& 3.0510& 0.5712 &0.5413& 0.4896 &6.1522& 5.1545& 2.9451 
	
	\\
	Const-5
	&3.3611& 3.0026& 0.5901 &0.5874& 0.5211 &5.0215& 4.9436& 3.8421
	
	\\
	Ada-0.1
	&3.0720& 2.9496& 0.6282 &0.5746& 0.5344 &4.2012& 4.7315& 3.6762 
	\\
	\rowcolor[RGB]{235,235,250}
	Ada-0.3
	&0.4645& 0.3005& 0.8705 &0.8375& 0.7705 &1.2617& 1.1709& 1.0667 
	\\
	Ada-0.5
	&1.7434& 1.1553& 0.7326 &0.7251& 0.6314 &2.2015& 2.0214& 1.5141 
	\\
	\bottomrule[2pt]
\end{tabular}

	}
	\label{tab:ablation_study_label_strategy}
\end{table}

\begin{table}[!t]
\centering
	\caption{Quantitative comparison between joint and separate training for anode and cathode plates.} 
   \resizebox{\linewidth}{!}{
    \setlength\tabcolsep{2pt}
    \renewcommand\arraystretch{1.2}
  \begin{tabular}{c||c|c|c|c|c|c|c|c}
  	\toprule[2pt]
  	Method  & AN-MAE$\downarrow$ 	& CN-MAE$\downarrow$ & AN-ACC$\uparrow$ & CN-ACC$\uparrow$ & PN-ACC$\uparrow$ & AL-MAE$\downarrow$  &CL-MAE$\downarrow$&OH-MAE$\downarrow$
  	\\
  	\hline

  	Joint Training
  	&0.4645& 0.3005& 0.8705 &0.8375& 0.7705 &1.2617& 1.1709& 1.0667 
  	\\
  	Separate Training
  	&0.4898 &0.3321 &0.8604 &0.8012 &0.7626 &1.2812 &1.1692 &1.1032 
  	\\
  	\bottomrule[2pt]
  \end{tabular}
  
	}
	\label{tab:ablation_study_two_channel}
\end{table}

\textbf{Joint Training vs. Separate Training.} 
Joint training refers to using a single unified model to simultaneously generate predictions related to both anode and cathode plates in a single training process, whereas separate training involves two independent models. 
Due to the use of a single set of parameters and a joint training process, a mutual learning mechanism can be established between anode and cathode, which helps enhance the model's discriminative ability and prevents confusion between the two plate types caused by the short overhang.
As shown in Tab.~\ref{tab:ablation_study_two_channel}, the jointly trained model achieves superior performance while using only half the parameters compared to the separate training setting. 

\begin{figure}[t]
	\centering
	\includegraphics[width=\linewidth]{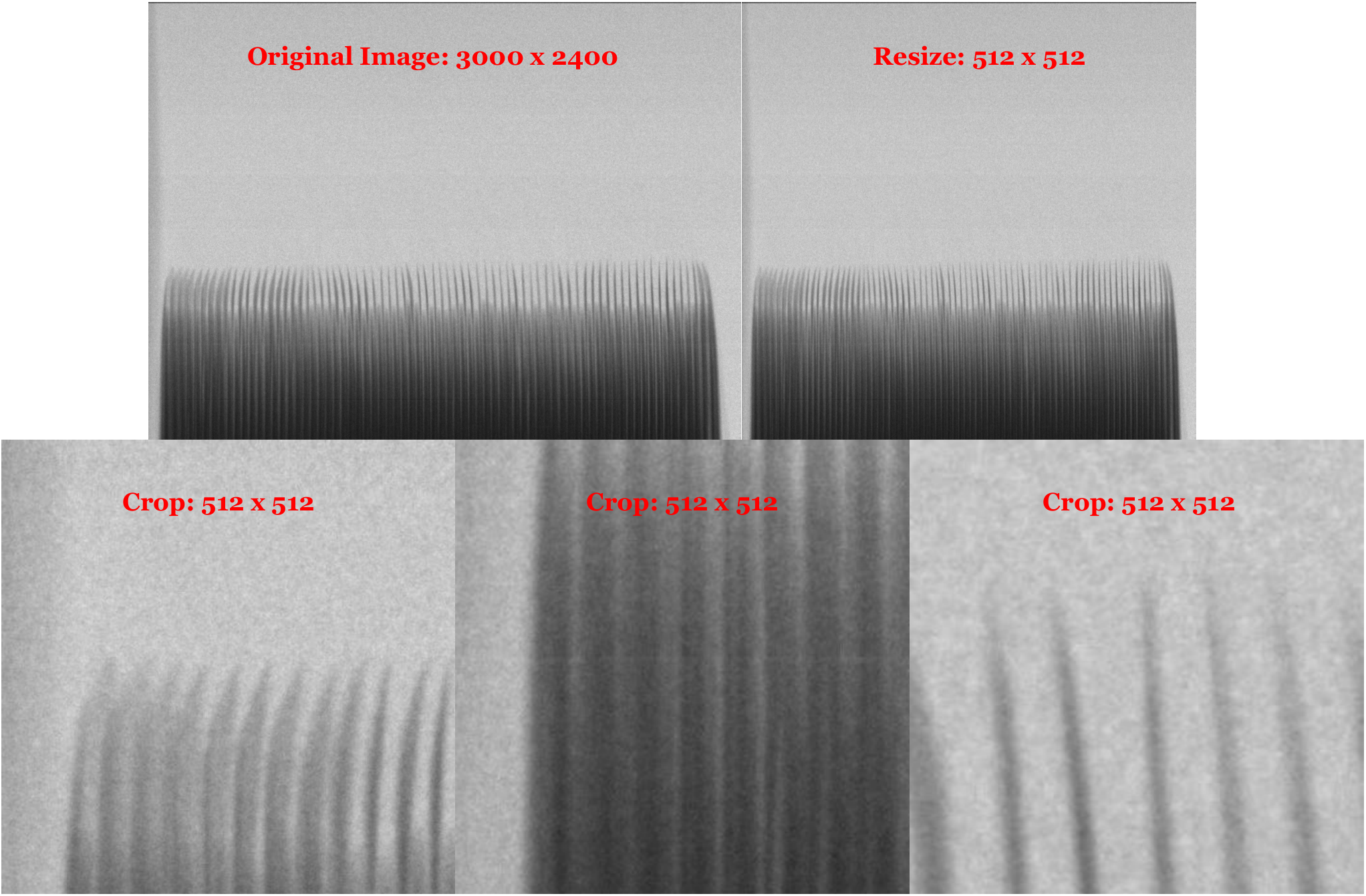}
	\caption{Viusal comparison between global resizing and patch-based cropping operation.}
	\label{fig:resize_vs_crop}
\end{figure}

\begin{table}[t]
	\centering
	\caption{Quantitative comparison between global resizing and cropping-based sliding-window strategies. The results are averaged across the three test set splits.}
	\resizebox{\linewidth}{!}{
		\setlength\tabcolsep{2pt}
		\renewcommand\arraystretch{1.2}
		\begin{tabular}{c||c|c|c|c|c|c|c|c}
			\toprule[2pt]
			Operation  & AN-MAE$\downarrow$ 	& CN-MAE$\downarrow$ & AN-ACC$\uparrow$ & CN-ACC$\uparrow$ & PN-ACC$\uparrow$ & AL-MAE$\downarrow$  &CL-MAE$\downarrow$&OH-MAE$\downarrow$
			\\
			\hline
			\rowcolor[RGB]{235,235,250}
			Resize
			&0.4645& 0.3005& 0.8705 &0.8375& 0.7705 &1.2617& 1.1709& 1.0667 
			\\
			Crop
			& 2.8457  &2.2754  &0.6854  &0.6547  &0.5787  &1.1573  &1.2024 & 1.1141
			\\
			\bottomrule[2pt]
		\end{tabular}
		
	}
	\label{tab:ablation_study_resize_crop}
\end{table}

\textbf{Effectiveness of Resizing Operation.} 
{
In this paper, we adopt a unified resizing strategy to 512$\times$512 for all experiments to ensure consistent training and fair comparison across methods. 
Directly processing full-resolution X-ray images (e.g., 3K$\times$4K) is impractical in real-world deployment scenarios due to its excessive GPU memory and inefficient runtime.
A commonly adopted alternative is tiled high-resolution inference using sliding windows.
However, as shown in Fig.~\ref{fig:resize_vs_crop}, this strategy introduces a critical limitation: \textit{electrode plates are frequently cut at tile boundaries, leading to fragmented endpoints and increased ambiguity in plate continuity}.
To quantitatively examine this effect, we conduct experiments using a pure cropping-based strategy, where models are trained on fixed-size 512$\times$512 crops without global resizing and evaluated using sliding-window inference at test time. As shown in Tab.~\ref{tab:ablation_study_resize_crop}, this tiled high-resolution setting consistently leads to noticeable performance degradation compared to global resizing. These results suggest that, for dense and highly structured plate layouts, preserving global structural continuity is more critical than maintaining local pixel-level resolution alone.
While resizing inevitably introduces geometric distortion, this is a standard and necessary practice in dense prediction tasks. Importantly, resizing does not eliminate endpoint structures, but represents them at a different scale. As shown in Fig.~\ref{fig:resize_vs_crop}, the relative geometric configuration of plate endpoints is preserved after resizing, allowing the network to learn consistent structural cues. Moreover, resizing acts as a form of implicit regularization by exposing the model to scale variations, which helps prevent overfitting to specific resolution patterns. 
Finally, we think high-fidelity modeling under ultra-high resolution belongs to a complementary research direction and often requires specialized designs, such as local-global fusion, multi-view learning, or hierarchical refinement strategies (e.g., HQ-SAM~\cite{HQSAM}, MVANet~\cite{MVANet}). Exploring such high-resolution or patch-augmented extensions in combination with MDCNeXt is a promising direction, which we explicitly identify as future work.}

\section{Future Works and Outlook}
This paper establishes a benchmark dataset and baseline for instance-level plate detection in power battery X-ray imagery. Despite promising results, many research directions and community challenges remain open for future exploration.

\textit{\textbf{\uppercase\expandafter{\romannumeral1}) Toward High-precision Segmentation and Unified Clue Modeling.}}
Our extensive experiments show that segmentation precision is strongly correlated with detection accuracy, especially on tough samples.
Thus, highly accurate segmentation will remain a key research focus. Beyond segmentation quality, holistic strategies that incorporate complementary clues such as bounding boxes, geometric layouts, and topological structures remain underexplored. For example, bounding boxes are more robust to overlap and adhesion, while segmentation excels at spatial localization. A promising direction is to design unified frameworks that fuse the strengths of both representations via selection, reweighting, or joint optimization.

\textit{\textbf{\uppercase\expandafter{\romannumeral2}) Blind Image Enhancement under Domain Constraints.}}
{
As shown in Fig.~\ref{fig:application_blind_image_enhance}, it highlights a practical imaging issue commonly encountered in industrial X-ray inspection. When the acquisition frame rate increases, reduced exposure time inevitably leads to motion blur, noise amplification, and ghosting artifacts. For MDCNeXt, such image degradation mainly affects fine boundary and endpoint localization, as blurred edges reduce local contrast, and also interferes with density-aware long-range modeling in densely packed plate regions, where noise disrupts stable state propagation. These effects are most evident in difficult samples with high plate density and severe overlap.
In principle, physical solutions such as prolonged exposure, multi-frame stacking, increased X-ray dosage, or high-sensitivity detectors can significantly improve image clarity and reduce blur and ghosting. However, these approaches often introduce increased acquisition time, manual intervention, and high tuning costs, making them unsuitable for real-time industrial deployment.
To handle image quality variability in a practical manner, we advocate domain-aware blind image enhancement under domain constraints. This strategy aims to improve X-ray image quality without requiring clean reference images, by leveraging multi-domain adaptive transfer learning and hybrid-domain unsupervised techniques that integrate natural image denoising or super-resolution priors with real PBD data. Once effective blind enhancement is achieved, MDCNeXt can operate on more stable and higher-quality inputs across different frame rates, reducing sensitivity to noise and blur while maintaining real-time throughput. This approach enables robust live inspection without being constrained by harsh imaging conditions or extensive manual calibration. 
}

\begin{figure}[t]
	\centering
	\includegraphics[width=1.0\linewidth]{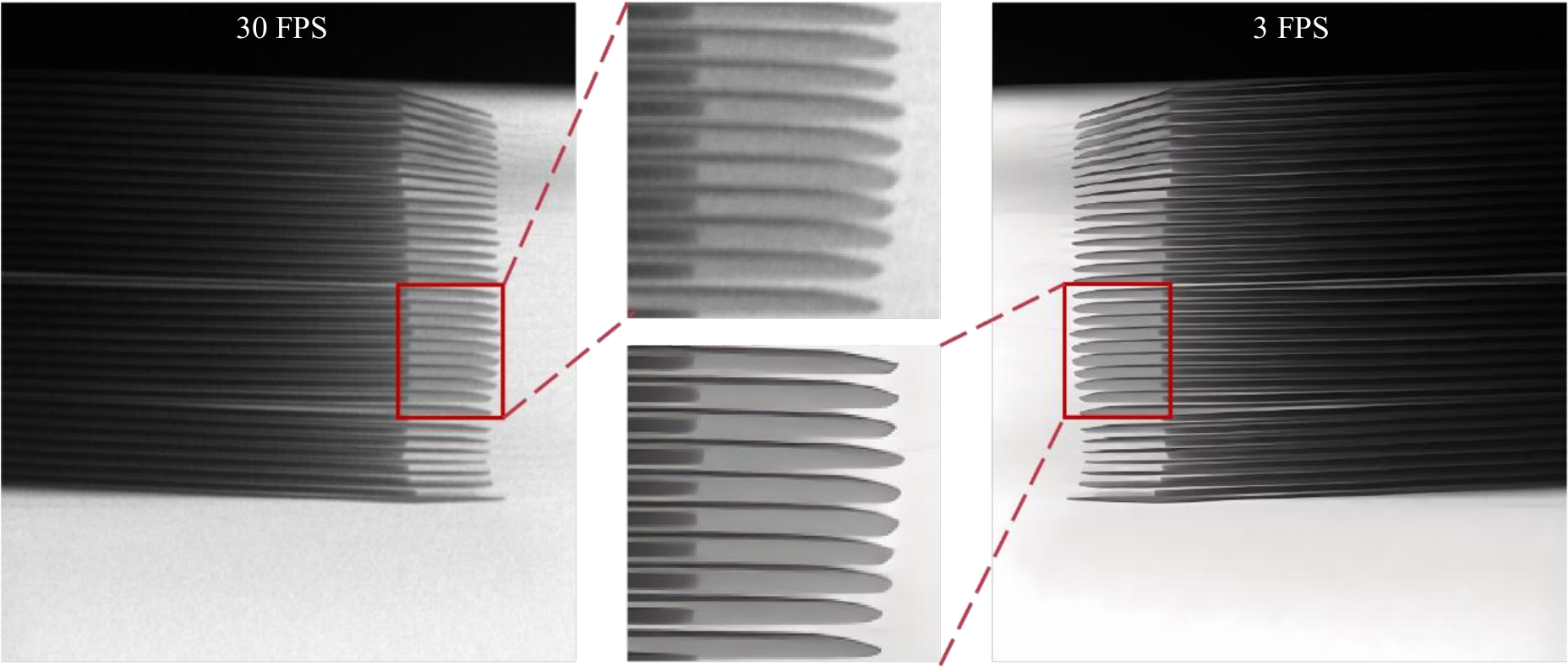}
	\caption{Comparison of X-ray image quality at 30 FPS and 3 FPS exposure times.}
	\label{fig:application_blind_image_enhance}
\end{figure}

\textit{\textbf{\uppercase\expandafter{\romannumeral3}) Fine-grained Industrial Image Generation.}}
High-quality and diverse battery data is essential for building robust foundation models in PBD. 
On the one hand, collecting real X-ray samples from different manufacturers is a long-term effort. On the other hand, we believe that generating targeted synthetic data based on specific interference attributes serves as an important complement, offering significant performance gains for PBD before sufficient real data is available. 
However, most general-purpose generative models~\cite{Stable_diffusion,ControlNet} are designed for natural scenes and focus on global content realism, lacking control over intricate local structures, such as individual defects or subtle texture changes. 
In the future, we plan to develop a controllable fine-grained generation framework tailored to industrial scenarios. Based on end-to-end training with industrial data, the focus will be on achieving pixel-level generation conditioned on interference types and spatial layouts, rather than merely ensuring local and global semantic consistency.

\begin{figure}[t]
\centering
\includegraphics[width=1.0\linewidth]{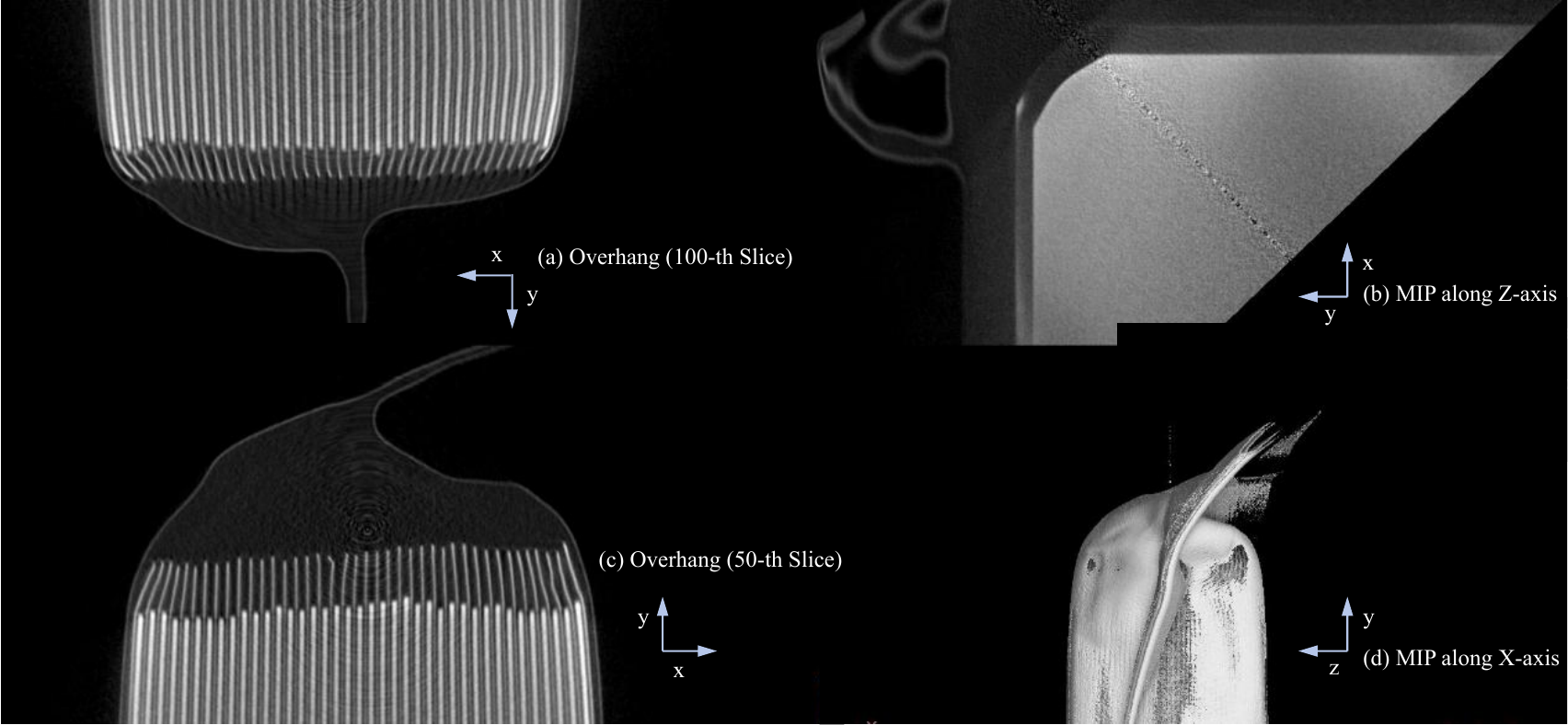}
\caption{Visualization of industrial CT data for battery inspection, including slice views and MIP images along different axes.}
\label{fig:application_CT_re}
\end{figure}

\textit{\textbf{\uppercase\expandafter{\romannumeral4}) Lifelong and Rapid Incremental Adaptation.}}  
Battery product iterations in industry often occur at a weekly or even faster pace, requiring detection models to quickly adapt to new specifications with minimal performance degradation for old classes. Thus, it is critical to develop lifelong learning and incremental adaptation strategies that support fast tuning with minimal labeled data, while avoiding catastrophic forgetting.

\textit{\textbf{\uppercase\expandafter{\romannumeral5}) High-speed and High-precision Industrial CT Reconstruction.}}  
As shown in Fig.~\ref{fig:application_CT_re}, we visualize  some  industrial CT data, including slice views for overhang inspection (a, c) and maximum intensity projection (MIP) images along different axes (b: Z-axis, d: X-axis) for metal inclusions or microcracks.
Compared with 2D X-ray imaging, volumetric CT provides more comprehensive structural insights, making it valuable for full-scope battery inspection.  
We plan to extend the PBD dataset into a 3D version by incorporating CT scans, which will significantly enhance the detection of internal quality issues. 
However, achieving real-time, high-resolution ($>$3K) CT reconstruction under industrial constraints remains highly challenging. 
Recent advances like 3D gaussian splatting (GS)~\cite{3DGS} have shown promise in surface-level reconstruction using multi-view inputs~\cite{Splatflow,VGGT}. 
Future research may explore hybrid neural-physical frameworks that incorporate CT physical projection models and 3D GS representations to enable end-to-end CT reconstruction from arbitrary views. In this process, sparse-view sampling, coarse-to-fine refinement pipelines, and the deployment of customized CUDA-based implementations can help accelerate the reconstruction speed.

\section{Conclusion}
We have presented the first comprehensive study on power battery detection (PBD) from complex industrial X-ray imagery. 
Specifically, we build a complex PBD5K dataset, formulate evaluation metrics and 
develop a segmentation-based multi-dimensional collaborative framework (\textit{i.e.}, MDCNeXt).
Compared with many potential modeling solutions and cutting-edge segmentation methods, our MDCNeXt demonstrates better performance under different  metrics and shows strong robustness in tough samples with dense plates and structural interference. The above contributions offer the community an opportunity to design new models for the PBD task. 
In the future, we plan to extend PBD5K to 3D CT data and encourage research on  domain-aware blind image enhancement, fine-grained industrial image generation, and high-speed and high-precision industrial CT reconstruction, toward more reliable full-scope battery inspection. 

\noindent\textbf{Data availability statement:} All datasets used and studied in this paper are publicly available.  The source code and datasets are publicly available at \href{https://github.com/Xiaoqi-Zhao-DLUT/X-ray-PBD}{PBD5K}.

\bibliographystyle{abbrv}
\bibliography{main}

\end{document}